\documentclass[10pt,journal,compsoc]{IEEEtran}
\ifCLASSOPTIONcompsoc
  \usepackage[nocompress]{cite}
\else
  \usepackage{cite}
\fi
\ifCLASSINFOpdf
\else
\fi
\usepackage{amsmath,amsfonts}
\usepackage{algorithmic}
\usepackage{algorithm}
\usepackage{array}
\usepackage[caption=false,font=normalsize,labelfont=sf,textfont=sf]{subfig}
\usepackage{textcomp}
\usepackage{stfloats}
\usepackage{url}
\usepackage{verbatim}
\usepackage{graphicx}
\usepackage{cite}
\usepackage{multirow}
\usepackage{verbatim}
\usepackage{amssymb}
\usepackage{threeparttable}
\usepackage{bbding}
\usepackage{color} 
\usepackage{diagbox}

\begin{document}

\title{Teacher Agent: A Knowledge Distillation-Free Framework for Rehearsal-based Video Incremental Learning}
%

%

\author{Shengqin Jiang,
	    Yaoyu Fang,
	    Haokui Zhang,
	    Qingshan Liu~\IEEEmembership{Senior Member,~IEEE},  
	    Yuankai Qi, 
	    Yang Yang~\IEEEmembership{Senior Member,~IEEE},
	    Peng Wang
\IEEEcompsocitemizethanks{\IEEEcompsocthanksitem S. Jiang, Y. Fang, Q. Liu are with the School of Computer Science, Nanjing University of Information Science and Technology, Nanjing, 210044, China. 
\IEEEcompsocthanksitem H. Zhang is with Intellifusion, Shenzhen, China.
\IEEEcompsocthanksitem Y. Qi is with Australian Institute for Machine Learning, the University of Adelaide,	Adelaide, SA 5005, Australia.
\IEEEcompsocthanksitem Y. Yang and P. Wang are with School of Computer Science and Engineering, University of Electronic Science and Technology of China, Chengdu, China.}
\thanks{Manuscript received August *, 2023; revised * *, 2023.   (\it{Corresponding author: Q. Liu, P. Wang.})}} 


\markboth{IEEE Transactions on Pattern Analysis and Machine Intelligence,~Vol.~*, No.~*, August~2023}%
{Shell \MakeLowercase{\textit{et al.}}: Bare Demo of IEEEtran.cls for Computer Society Journals}
%


\IEEEtitleabstractindextext{%
\begin{abstract}
Rehearsal-based video incremental learning often employs knowledge distillation to mitigate catastrophic forgetting of previously learned data. However, applying this approach directly to video incremental tasks faces two major challenges. Firstly, learning video sequences alone requires substantial computing resources, and loading the network from the previous stage for experience replay further exacerbates the burden. Secondly, the knowledge review capability of the network is constrained by the performance of the teacher network from the previous stage. We revisit this issue and empirically confirm that inaccurate predictions by the teacher model for some memorized exemplars directly limit the performance of knowledge review.  To address these problems, we first propose a knowledge distillation-free framework for rehearsal-based video incremental learning called \textit{Teacher Agent}. Instead of loading parameter-heavy teacher networks, we introduce an agent generator that is either parameter-free or uses only a few parameters to obtain accurate and reliable soft labels. This method not only greatly reduces the computing requirement but also circumvents the problem of knowledge misleading caused by inaccurate predictions of the teacher model. Moreover, we put forward a self-correction loss which provides an effective regularization signal for the review of old knowledge, which in turn alleviates the problem of catastrophic forgetting. Further, to ensure that the samples in the memory buffer are memory-efficient and representative, we introduce a unified sampler for rehearsal-based video incremental learning to mine fixed-length key video frames. Interestingly, based on the proposed strategies, the network exhibits a high level of robustness against spatial resolution reduction when compared to the baseline. Extensive experiments demonstrate the advantages of our method, yielding significant performance improvements while utilizing only half the spatial resolution of video clips as network inputs in the incremental phases. 
\end{abstract}

}

\maketitle

\IEEEdisplaynontitleabstractindextext

%
\IEEEpeerreviewmaketitle

\IEEEraisesectionheading{\section{Introduction}\label{sec:introduction}}

\IEEEPARstart{O}{ver} the past few years, the proliferation of short-form video apps like Instagram and TikTok has resulted in an unprecedented surge in multimedia data, particularly videos. Effectively comprehending and analyzing these videos can facilitate more efficient management and offer better services to users. While deep learning has made substantial progress in action recognition, most existing methods focus on predetermined categories. With the emergence of fresh themes and trends, novel video categories are emerging, rendering previous recognition methods inadequate.

\begin{figure}[htbp]
	\centering
	\includegraphics[scale=0.7]{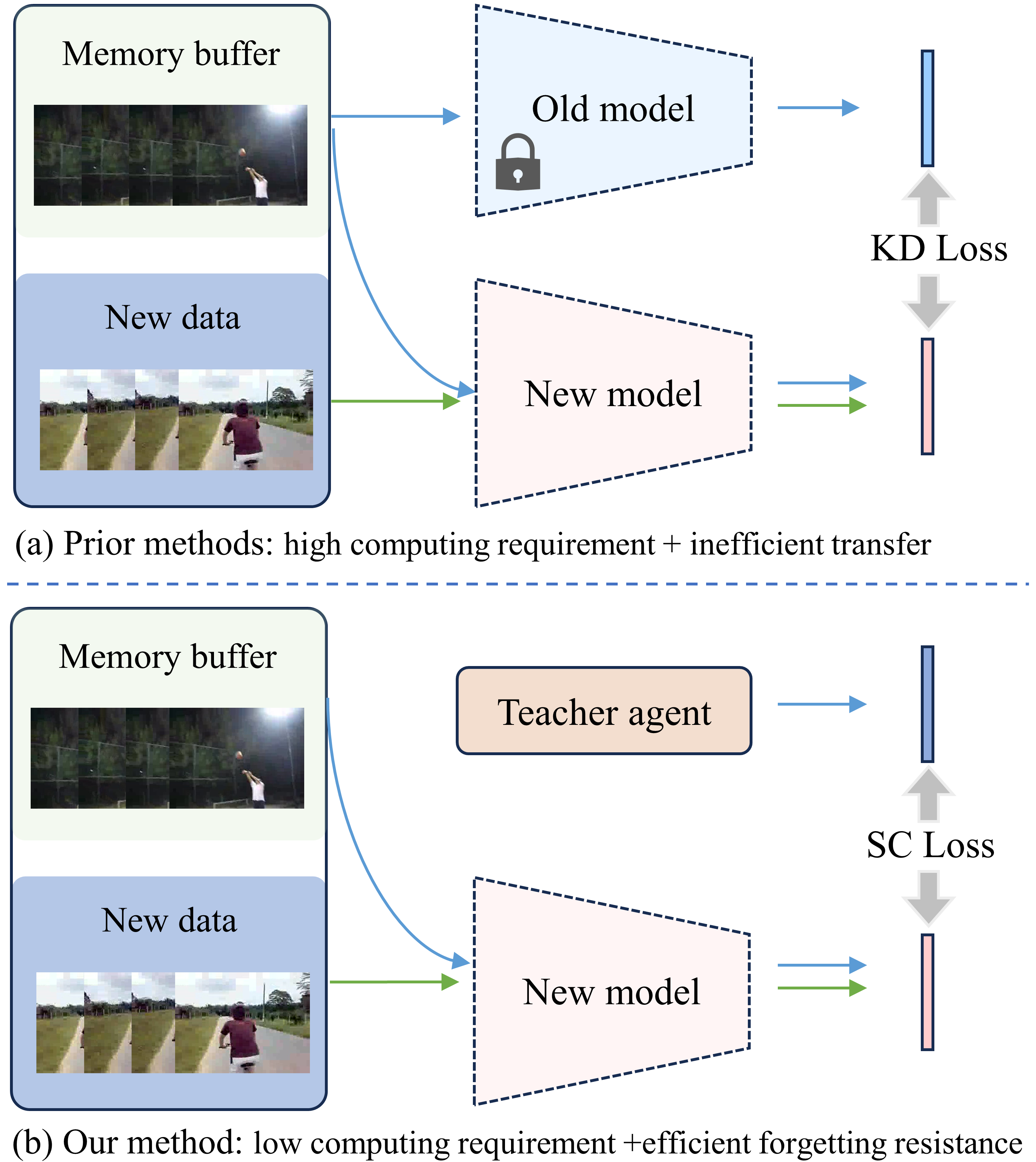}
	\caption{Comparison of rehearsal-based methods and our method. Rehearsal-based methods rely on the network trained in the previous stage for knowledge distillation, helping the current model review old knowledge. However, for action recognition, these methods impose a substantial computational overhead, and when using teacher models with limited performance, the knowledge transfer may not be as effective. In contrast, our approach introduces a teacher agent that eliminates the need to load a pre-trained model and can efficiently resist catastrophic forgetting.}
	\label{compare}
\end{figure}

In this scenario, repeatedly training an action recognition model from scratch whenever new video classes emerge is impractical due to the significant computational overhead and laborious nature of the process. Additionally, in practice, it may not be feasible to have access to all the old samples due to safety considerations. One potential solution is to fine-tune the existing models using videos of the new classes. However, this may lead to the catastrophic forgetting problem, whereby the model achieves excellent performance on the new video classes but experiences a significant drop in performance on the old ones. Therefore, it is essential for the recognition model to adapt to the new classes while retaining the critical knowledge learned from past data.\par

Another feasible solution to overcome the problem of forgetting old knowledge is to learn new classes incrementally. Class Incremental Learning (CIL) is a common method used for this purpose. iCaRL is a representative method that maintains past knowledge by preserving some representative samples, known as exemplars~\cite{rebuffi2017icarl}, and using knowledge distillation to avoid catastrophic forgetting. Recently, Park et al.~\cite{park2021class}  tackled the challenge of CIL for video action recognition by employing a straightforward frame-based feature representation approach to store exemplars from previously learned tasks. In the same year, Zhao et al.~\cite{zhao2021video} decomposed spatio-temporal knowledge before distillation, rather than treating it as a whole during knowledge transfer. More recently, Villa et al.~\cite{villa2022vclimb} introduced a new benchmark for video continual learning and proposed a temporal consistency regularization method to enforce representation consistency between downsampled and original samples. These studies have made significant contributions to the development of incremental learning in the video domain. It is worth noting that, as in the image domain, the issue of catastrophic forgetting has yet to be adequately addressed.


As discussed above, rehearsal-based methods are commonly used to address the issue of catastrophic forgetting by knowledge distillation~\cite{rebuffi2017icarl,park2021class,zhao2021video,villa2022vclimb}, as shown in Fig.~\ref{compare}. However, this method has to face two significant challenges. Firstly, the video classification task itself demands substantial computational resources, and incorporating the teacher network from the previous training stage for distillation further worsens the computational burden. Secondly, the performance of the teacher model  constrains the potential improvement in the current stage. This limitation stems from both the inherent performance constraints of the teacher network and the issue of catastrophic forgetting, which hinders accurate predictions for memorized samples. Empirical studies~\cite{zhao2020maintaining} also identified this problem and attributed it to the high bias in the last FC layer's weights. However, we argue that the primary cause is the teacher's inability to consistently provide accurate predictions. This challenge is further compounded by the limitations on the number of memorized samples in rehearsal methods, which make it difficult to train a strong teacher network.

To address this issue, we conducted empirical studies to revisit the impact of knowledge distillation on the network in the current stage. Our findings indicate that removing the distillation loss of incorrect samples results in a slight improvement in the network's performance. But this also leads to the reduced utilization rate of the selected exemplars, emphasizing the importance of correct prediction. To offer an alternative to knowledge distillation-based incremental learning, we introduce a teacher agent for experience replay. At the core of the teacher agent lies the teacher generator, responsible for consistently producing accurate and reliable soft labels. This generator can be designed with a minimal number of parameters or even none at all, thus avoiding the redundant computation of the teacher model. Building upon the labels, we introduce a new self-correction loss designed to facilitate the network in reviewing old knowledge.  Moreover, in order to enhance the memory efficiency and representativeness of samples in the memory buffer, we introduce a unified sampler for rehearsal-based video incremental learning that focuses on mining fixed-length key video frames. By adopting these proposed methods, our network shows increased robustness against resolution reduction compared to baseline. Impressively, our method performs favorably against recent state-of-the-art methods on four challenging datasets with only 26\% FLOPs of baseline.  We hope our findings will inspire researchers to explore this alternative method further instead of solely relying on different methods of knowledge distillation. The main contributions are summarized as follows:

\begin{itemize}
	\item We propose a knowledge distillation-free network framework for rehearsal-based video incremental learning. This enhances the network's ability to resist catastrophic forgetting without requiring high computational requirements.
	\item We put forward a teacher generator with no or few parameters to generate accurate and reliable soft labels. This avoids the huge computational overhead required to load the teacher network for knowledge distillation and the misclassification caused by the limited performance of teacher network.
	\item We propose a novel self-correction loss that serves as an effective regularization signal to facilitate the retention of previously acquired knowledge. 
	\item An explainable unified sampler is introduced to mine key-frames of the exemplars in the memory buffer, which is beneficial for efficient storage management. 
	\item Our method exhibits heightened robustness across inputs of varying spatial resolutions, surpassing the capabilities of the baseline method. Notably, even when operating at half the resolution, our method achieves a substantial performance improvement compared to some state-of-the-art methods.
\end{itemize}

\section{Related Work}

\subsection{Image Class Incremental Learning}

{In various practical applications, it is often essential to gradually learn new classes from streaming data, known as class incremental learning. Driven by this growing demand, numerous studies have concentrated on tasks related to image classification, engaging in comprehensive exploration. These efforts can be generally categorized into four distinct categories: architecture, parameter regularization, knowledge distillation, and rehearsal. Subsequently, we will provide a concise overview of these endeavors.} \par


Architecture-based methods are commonly employed to handle new tasks  through network extensions or modifications. One such example is the work by DER~\cite{yan2021dynamically}, where they introduced a dynamic expansion of representation method. This approach involves freezing the previously learned parameters and then extending the model with additional feature dimensions using a new mask-based feature extractor. Zhu et al.~\cite{zhu2022self} utilized a dynamic restructuring strategy, which creates structural space for learning new classes while keeping the old classes' space stable by maintaining heritage at the main branch and fusing updates at the side branch.

Parameter regularization methods typically estimate the importance of each model parameter by analyzing the sensitivity of the loss function to changes in the parameter. For example, Kirkpatrick et al.~\cite{kirkpatrick2017overcoming} employed Fisher's information matrix to identify the parameters important for the previous task, and then reduced the changes of these parameters when learning new tasks using a one-parameter canonical loss. Aljundi et al.~\cite{aljundi2018memory} developed a mechanism that identifies the most crucial weights in the model by examining the sensitivity of the output function, rather than the loss. While these methods can alleviate catastrophic forgetting to some extent, they may not achieve satisfactory performance under challenging settings or complex datasets.

Methods based on knowledge distillation encourage the model to learn new tasks by imitating the representation of the old model trained on the previous task~\cite{zhang2020class}. PODNet~ \cite{douillard2020podnet} was proposed to utilize space-based distillation losses to limit model changes. Cheraghian et al.~\cite{cheraghian2021semantic} introduced semantic-aware knowledge distillation, which uses word vectors as auxiliary information and constructs knowledge extraction terms to reduce the impact of catastrophic forgetting. Rehearsal-based methods are an extension of knowledge distillation methods, where a limited number of samples from previous tasks are stored in a memory buffer during the training of a new task.  For instance,  iCaRL~\cite{rebuffi2017icarl} was the first to use replay, where each class retains a small number of samples to approximate the class centroid. Another method, Rainbow~\cite{bang2021rainbow}, proposed a novel memory management strategy based on per-sample classification uncertainty and data augmentation. Similar to herding-based methods~\cite{rebuffi2017icarl,wu2019large,hou2019learning},  Liu et al.~\cite{liu2020mnemonics} modified the herding procedure by parameterizing exemplars, enabling optimization in an end-to-end manner. The aforementioned studies have significantly contributed to the advancement of incremental learning in image classification tasks. However, they have not sufficiently delved into the realm of spatio-temporal tasks, thus leaving room for further exploration in this area.

\subsection{Video Class Incremental Learning}

Action recognition is a crucial and highly demanding task of spatio-temporal analysis, garnering significant attention in both research and industry. Existing work predominantly centers on the static category, with comparatively fewer studies addressing incremental tasks. In what follows, we will provide a concise overview of the advancements made in both realms.

Action recognition within static categories primarily involves predicting these specific categories by learning from video sequences associated with fixed behaviors. The present study can be broadly categorized into two types based on the configuration of stacked layers within the network: CNN-based methods and transformer-based methods.  The former one typically uses 2D or 3D convolution layers to build an effective network architecture~\cite{carreira2017quo, li2020directional,wang2021tdn}. I3D~\cite{carreira2017quo} inflated the convolutional and pooling kernels of a 2D CNN with an additional temporal dimension. Wang et al.~\cite{wang2021action} introduced a multi-path excitation module that models spatial-temporal, channel-wise, and motion patterns separately. Transformer-based methods, on the other hand, exploit the ability of transformers to capture global information and model spatio-temporal relationships in video frames~\cite{arnab2021vivit,liu2022video}. Dosovitskiy et al.~\cite{dosovitskiy2020image} demonstrated that a pure transformer applied directly to sequences of image patches can perform well on image classification tasks. Following this idea, Yang et al.~\cite{yang2022recurring} proposed a recurrent vision transformer framework that leverages spatial-temporal representation learning to handle the video motion recognition task.

{To perceive dynamically changing behavioral categories, video class incremental learning offers an effective learning paradigm. Thus far, a small amount of work has been done in this direction. Park et al.~\cite{park2021class} studied time-channel importance maps and leveraged these maps to learn the representations of incoming examples via knowledge distillation. Zhao et al.\cite{zhao2021video} initially decomposed spatio-temporal features prior to knowledge transfer. They introduced a dual-granularity exemplar selection method that effectively identifies and stores representative video instances of old classes, along with key-frames within videos. Subsequently, Villa et al.\cite{villa2022vclimb} contributed to the field by introducing a standardized test-bed for video continual learning. This benchmark, known as vCLIMB, aims to analyze the phenomenon of catastrophic forgetting in deep models, specifically within the context of video continual learning. The aforementioned studies have significantly contributed to advancing the field. It is evident that knowledge distillation plays a crucial role in reviewing and consolidating past knowledge within these models. However, it's worth noting that integrating knowledge distillation involves loading the network acquired in the previous stage, leading to increased resource consumption in the already resource-intensive task of training spatio-temporal models. Furthermore, our empirical observations reveal that the incorrect reasoning of the teacher network can misguide the learning process of the current stage network. These issues serve as the primary motivation for the current work.}

{More recently, some efforts to advance this field have been notable, with PIVOT~\cite{villa2023pivot} serving as a representative example. PIVOT utilized the CLIP model, pre-trained on an extensive dataset, as its backbone network. Through fine-tuning the newly added prompts, it attains impressive performance. Note that our work differs from PIVOT in the following aspects. We focus on addressing the challenges posed by knowledge distillation for incremental learning, particularly the issues of misguidance and additional resource consumption. In contrast, PIVOT primarily concentrates on how to leverage the CLIP model pre-trained on large-scale image-text pairs for video incremental learning tasks. This means that the starting point and framework of the two works are different. Additionally, our work addresses video sequences downsampled at spatial resolution in the incremental stage,  a characteristic that renders them incompatible with PIVOT that only allows to receive fixed spatial inputs.}

\section{Preliminaries}
\subsection{Problem Formulation}

\noindent\textbf{Base task.} Typically, in the incremental learning setups ~\cite{kang2022class, zhou2022forward,aljundi2018memory,zhao2021video,zhao2021video}, the model $F_1(\cdot, \theta_1)$ is trained on a video dataset  ${\mathcal{D}_{1}} = \{ (I_i^1,y_i^1)\} _{i = 1}^{{M_1}}$, referred to as the base task or base session. Here, $I_i^1 \in \mathbb{R}^{t \times c \times h \times w}$ represents the $i$th training video sample with $t$ frames, $c$ channels, height $h$ and width $w$, and $y_i^1 \in \mathcal{C}_1$ is its corresponding label from the label space $\mathcal{C}_1$ of the base task. The objective of this session is to optimize the model $F_1(\cdot, \theta_1)$ by minimizing the following objective function:

\begin{equation}
	\frac{1}{{\left| {{\mathcal{D}_1}} \right|}}\sum\nolimits_{(I,y) \in {\mathcal{D}_1}} {\mathcal{L}({F_1}(I,{\theta _1}),y)}
\end{equation}
where $\mathcal{L}(\cdot)$ denotes the classification loss (cross-entropy loss (CE loss) by default). Without losing generality, the model $F_1(\cdot, \theta_1)$ is composed of a convolutional neural network, consisting of a feature extractor $\phi  = f(I,\omega_1)$ and a classifier $g(\phi, W_1) = {W_1^{\rm T}}\phi$, where ${\theta _1} = \{ {\omega _1},{W_1}\}$ denotes the network parameters. Here, $W_1 \in \mathbb{R}^{l \times |\mathcal{C}_1|}$, where $l$ is the dimensionality of the embedding vector $\phi \in \mathbb{R}^{l}$.

\noindent\textbf{Incremental task.} In the incremental task,  $\mathcal{D}_{2}, \mathcal{D}_{3}$ and subsequent video datasets arrive in order, each containing one or more new classes. Specifically, for the $\kappa$-th task, the dataset {$\mathcal{D}_{\kappa}$}  includes  ${M}_{\kappa}$ training samples with {$|\mathcal{C}_{\kappa}|$} new classes, denoted as ${\mathcal{D}_\kappa} = \{ (I_i^\kappa, y_i^\kappa)\} _{i = 1}^{{M_\kappa}}$, where  ${I}_{i}^{\kappa}$ represents the $i$th video sample, and $y_i^\kappa$ $\in$ $\mathcal{C}_{\kappa}$ is its corresponding label. The label space of task $\kappa$ is defined as $\mathcal{C}_\kappa$. It is important to note that all label sets are mutually exclusive, i.e., $\mathcal{C}_{i}$ $\cap$ $\mathcal{C}_{j}$ = $\emptyset$ ($i \ne j$) for $\forall {i}, {j} > 1$.

In this paper, we introduce a general incremental setup for video recognition, namely, \textit{downsampling-resolution incremental learning}. Specifically, in the $\kappa$th incremental session, we represent each video sample as $I^\kappa \in {R^{t \times c \times (\delta h) \times (\delta w)}}$, where $0 < \delta \leqslant 1$ is the scaling factor of spatial resolution. The motivation of this setting is two-fold: (1) In real-world applications, newly added video categories may come from diverse scenes, resulting in videos with varying spatial resolutions. (2) Video classification models inherently exhibit high computational complexity, with spatial resolution being a significant factor contributing to this burden. Hence, we standardize the inputs of different resolutions into a unified downsampling format during incremental stages. By doing so, we significantly reduce computational overhead and  facilitate the deployment of resource-limited devices. This incremental task is challenging because the loss of details in video frames increases the difficulty of network learning.



\subsection{Incremental Learning with Knowledge Distillation}

Class-incremental learning for video recognition targets training a single model effectively, enabling it to continuously learn new classes while avoiding forgetting previously learned classes. Knowledge distillation is a common and crucial technology used in the majority of previous works to do this~\cite{rebuffi2017icarl,villa2022vclimb}. It uses the network model trained in the previous stage as the teacher network to guide the network in the current stage in reviewing old knowledge. The optimization objective is formulated as follows:

\begin{equation}
	\begin{gathered}
		\frac{1}{{\left| {{\mathcal{D}_\kappa }} \right|}}\sum\nolimits_{(I,y) \in {\mathcal{D}_\kappa }} {\mathcal{L}({F_\kappa }(I,{\theta _\kappa }),y)}  +   \hfill \\
		\frac{1}{{\left| {{\mathcal{P}_{\kappa  - 1}}} \right|}}\sum\nolimits_{(I,y) \in {\mathcal{P}_{\kappa  - 1}}} {\mathcal{K}({F_\kappa }(I,{\theta _\kappa }),y)} \hfill \\ 
	\end{gathered} 
\end{equation}
where  $\mathcal{P}_{\kappa - 1}$ refers to the memorized samples from the previously learned tasks up to the ($\kappa-1$)th task, and $\mathcal{K}(\cdot)$ denotes the distillation loss. To transfer knowledge, CE loss and knowledge distillation loss (KD loss) are often employed to optimize the network:

\begin{equation}\label{eq3}
	\begin{aligned}
		\mathcal{K}({F_\kappa }(I,{\theta _\kappa }),y) = \lambda {L_{ce}}(I;{\theta _\kappa }) + (1 - \lambda ){L_{kd}}(I;{\theta _\kappa }),
	\end{aligned}
\end{equation}
where $\lambda$ is a hyper-parameter ($0.5$ by default). The KD loss is given by~\cite{villa2022vclimb}:

\begin{equation}
	\begin{aligned}
		{L_{kd}}(I;{\theta _\kappa }) = - \hat q(I)log(q(I)),
	\end{aligned}
\end{equation}
where  ${\hat{q}(I)}$ refers to the soft label of ${I}$ obtained from the teacher model, and ${q(I)}$ represents the softened probability of the network output.

\DeclareGraphicsExtensions{.pdf,.jpeg,.png,.jpg}
\begin{figure}[htbp]
	\centering
	\includegraphics[scale=0.62]{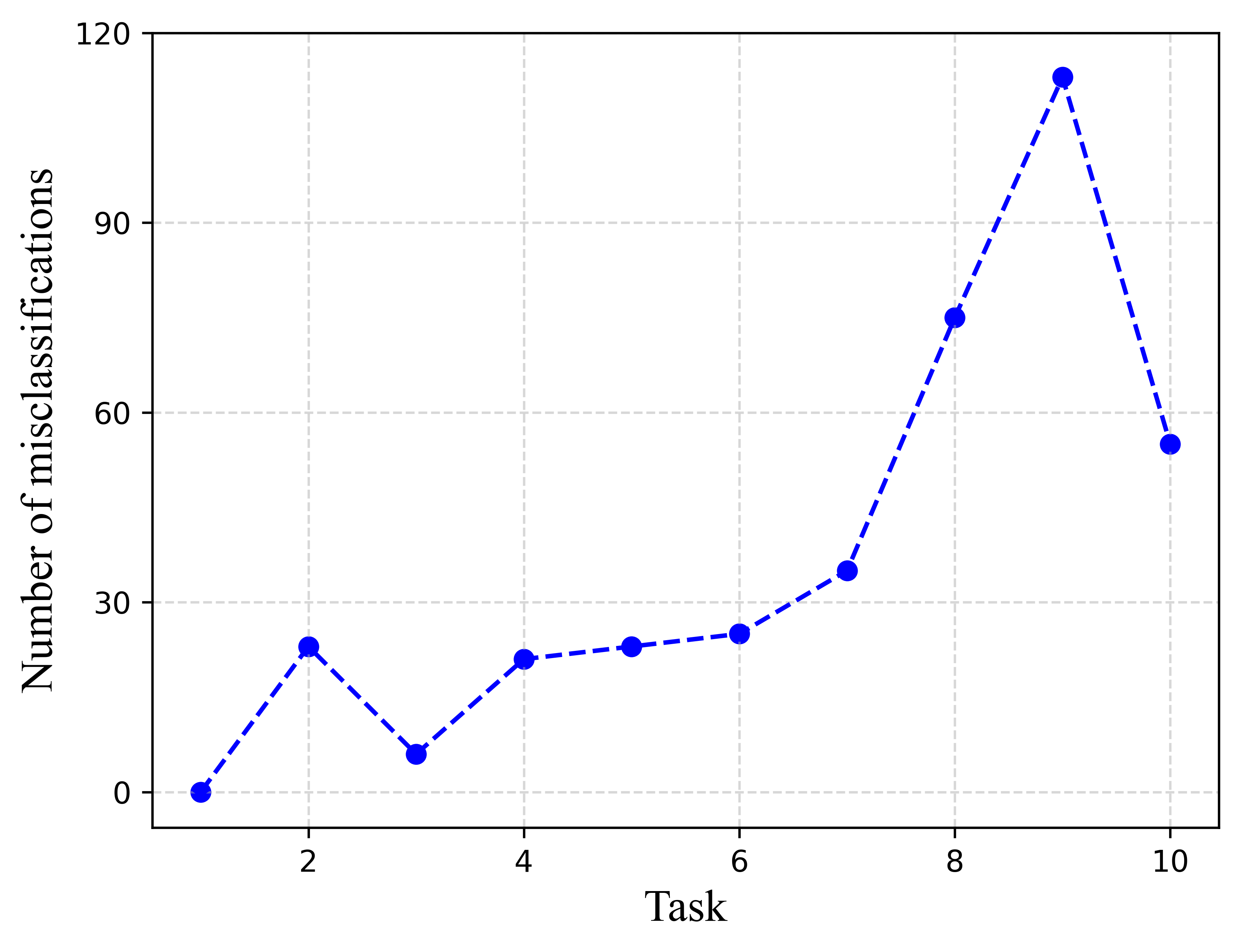}
	\caption{{Number of misclassifications from teacher network as the task progresses. With an increase in the number of tasks, there is a corresponding rise in the number of misclassified memory exemplars.}}
	\label{wrong_num}
\end{figure}

As discussed previously, knowledge distillation suffers from two drawbacks: high computational complexity and the limited guidance of  teacher model. To explore the impact of the latter, we conducted experiments using the incremental network proposed in~\cite{villa2022vclimb} as our baseline under the incremental setup with a hyperparameter value of $\delta=0.5$. As shown in Fig.~\ref{wrong_num}, the results reveal that, even when exemplars are selected using herding, there still remain prediction biases in the teacher network for a certain amount of the memorized exemplars. As the number of tasks increases, catastrophic forgetting of the previously learned classes becomes increasingly severe. These biased predictions result in the accumulation of deviations in the student model. While a recent study suggests that a poorly-trained teacher with lower accuracy can still improve the student's performance, the positive impact is limited~\cite{yuan2020revisiting,zhao2020maintaining}.

\DeclareGraphicsExtensions{.pdf,.jpeg,.png,.jpg}
\begin{figure*}[htbp]
	\centering
	\includegraphics[scale=0.75]{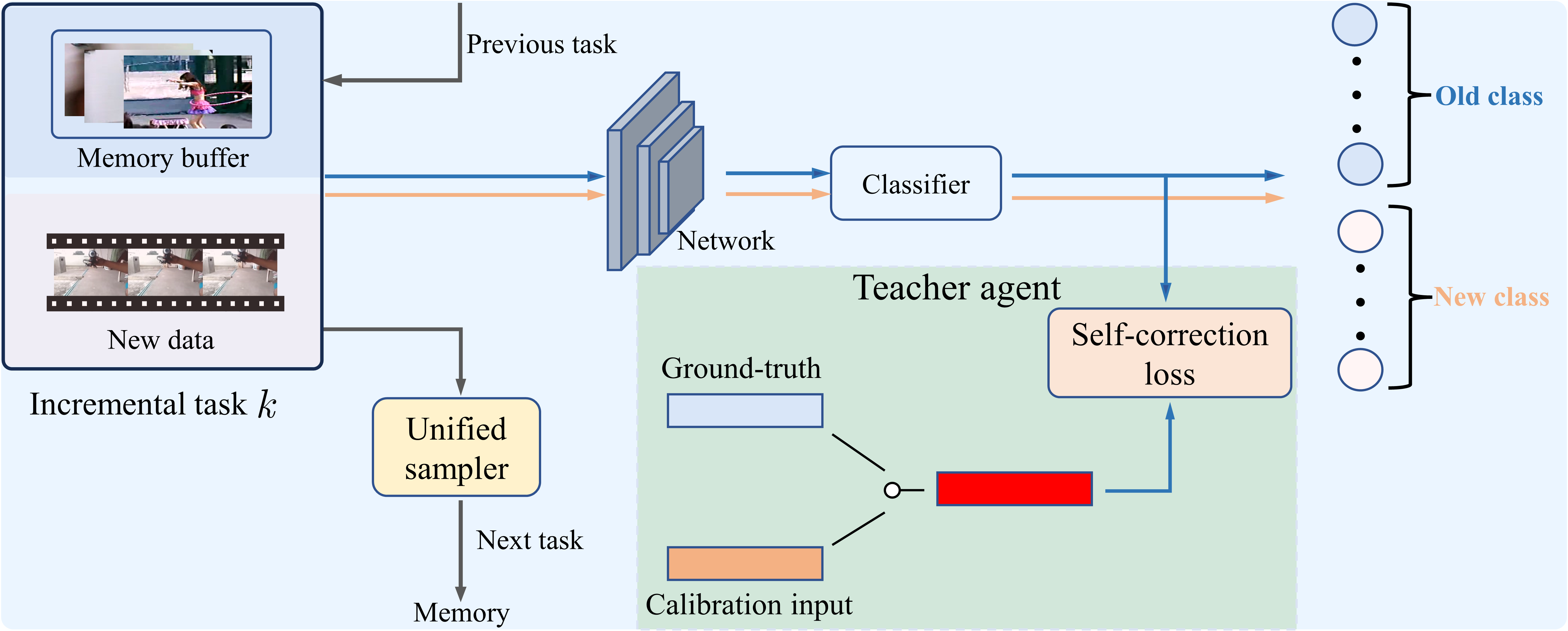}
	\caption{Framework overview of our proposed method. We propose a teacher agent-based incremental framework for video recognition instead of distillation-based incremental ones. It contains a teacher agent for experience replay and a unified sampler for fixed-length key-frame sampling. The teacher agent consists of two key designs: a teacher generator to generate trustworthy soft labels, and a self-correction loss to optimize the network parameters and thus help the network consolidate old knowledge. The proposed strategies demonstrate efficient computational efficiency and good resistance to forgetting, even in the face of downsampling-resolution incremental settings.}
	\label{network}
\end{figure*}

A straightforward method to avoid the accumulation of deviations is to employ on-off control and Eliminate gradient back-propagation of Misclassified Exemplars, short for EME. As shown in the first and second rows of Table~\ref{scloss} (Sec.~\ref{exper}), baseline (EME) performs slightly better than baseline, despite the fact that not all the samples stored in the episodic memory are utilized for knowledge distillation. These results highlight the crucial importance of accurate soft predictions in preserving old knowledge. In other words, a stronger teacher is required to provide effective assistance in reviewing prior knowledge. However, the rehearsal-based methods assume that the memory buffer or the number of samples is limited, which can be problematic when dealing with the loss of previous class knowledge brought about by continual learning of incremental tasks. It is worth noting that even if we can mitigate the effects of misclassified predictions, we are still not making full use of the exemplars stored in the memory.

Based on the analysis above, we argue that the key criterion for a successful teacher in the knowledge review process is the ability to make consistently accurate predictions as much as possible. Nevertheless, training such a teacher can be costly due to limited memory buffers. To this end, we propose an efficient solution to achieve better knowledge review, which will be detailed in the next section. 



\section{Method}

\textit{Building upon the discussions above, we challenge the conventional distillation paradigm used in rehearsal-based methods and propose a teacher agent instead.} This strategy avoids the huge computational overhead associated with loading teacher models while circumventing the limitations imposed by limited teacher performance in knowledge distillation. It consists of two key designs: teacher generator and Self-Correction (SC) loss. Moreover, we introduce a unified sampler mining a fixed number of discriminative video frames. This helps the model to better recall old knowledge in addition to saving storage costs. The overview of our network is shown in Fig.~\ref{network}. Next, we'll go over each of them in detail. 

\subsection{Teacher Agent}

\textbf{Teacher generator.} It is introduced to produce a soft label that is positively correlated with the ground-truth of the old classes, with the aim of replacing the output of the teacher network. {Specifically, the ground-truth of the exemplar $x$ in episodic memory is denoted by a one-hot vector ${{\hat y}^\kappa } \in {R^{{\hbar _\kappa }}}$  for the $\kappa$-th task, where ${\hat y}^{\kappa}(c)$ = 1 and ${\hat y}^{\kappa}(m)$ = 0$|_{c, m\in[1,{\hbar _\kappa }],m \neq c}$ for the $c$-th class, ${\hbar _\kappa } = \sum\limits_{i = 1}^\kappa  {|{\mathcal{C}_i}|}$, and $|{\mathcal{C}_i}|$ denotes the number of categories at $i$-th task.} The generator outputs new soft labels using the following operations:

\begin{equation}\label{eq5}
	\begin{aligned}
		\chi_{\kappa}(x)=\frac{{({\hat y}^{\kappa}+p^{\kappa})}^{\alpha}}{\sum{({\hat y}^{\kappa}+p^{\kappa})}^{\alpha}}
	\end{aligned}
\end{equation}
where  $\alpha  \in (0,1]$ is a regulating factor, smoothing each element of summed vectors. ${p^\kappa } \in {R^{{\hbar _\kappa }}}$ denotes a calibration input vector by normalization, and each element of $p^{\kappa} $  falls within the range of 0 and 1.  We utilize the sigmoid activation function to implement the normalization. The primary objective of this design is to mimic the teacher network, thereby generating robust soft labels while preventing overconfidence of labels. The calibration input provides a high level of flexibility and can be configured with few or no parameters, depending on specific requirements. Here, we briefly discuss two feasible options:

\textit{Parameter-Free Calibration Input}: This option involves generating a calibration input using a specific distribution such as uniform distribution. The advantages of this method are two-fold: it minimizes the consumption of computing resources and simultaneously enhances label diversity. This strategy increases the diversity of calibration values, preventing the final generated soft labels from getting stuck in a local value, thereby allowing the network to obtain more possibilities for feature expression.

\textit{Parameterized Calibration Input}: Alternatively, a version with learnable parameters can be utilized to generate a series of calibrations. {Specifically, the calibration input vector $p^{\kappa}$ in Eq.~(\ref{eq5}) is generated by a learnable parameter vector.} This design allows for better adaptability through gradient updates, thus enabling the input to obtain optimal parameters.

Interestingly, our experiments have demonstrated that both of these choices can significantly improve performance. Moreover, it is worth noting that the calibration strategy exhibits robustness, showing less sensitivity to the specific choices (refer to Table~\ref{scloss} (Sec.~\ref{exper}) for details).

\begin{figure}[htbp]
	\centering
	\includegraphics[scale=0.57]{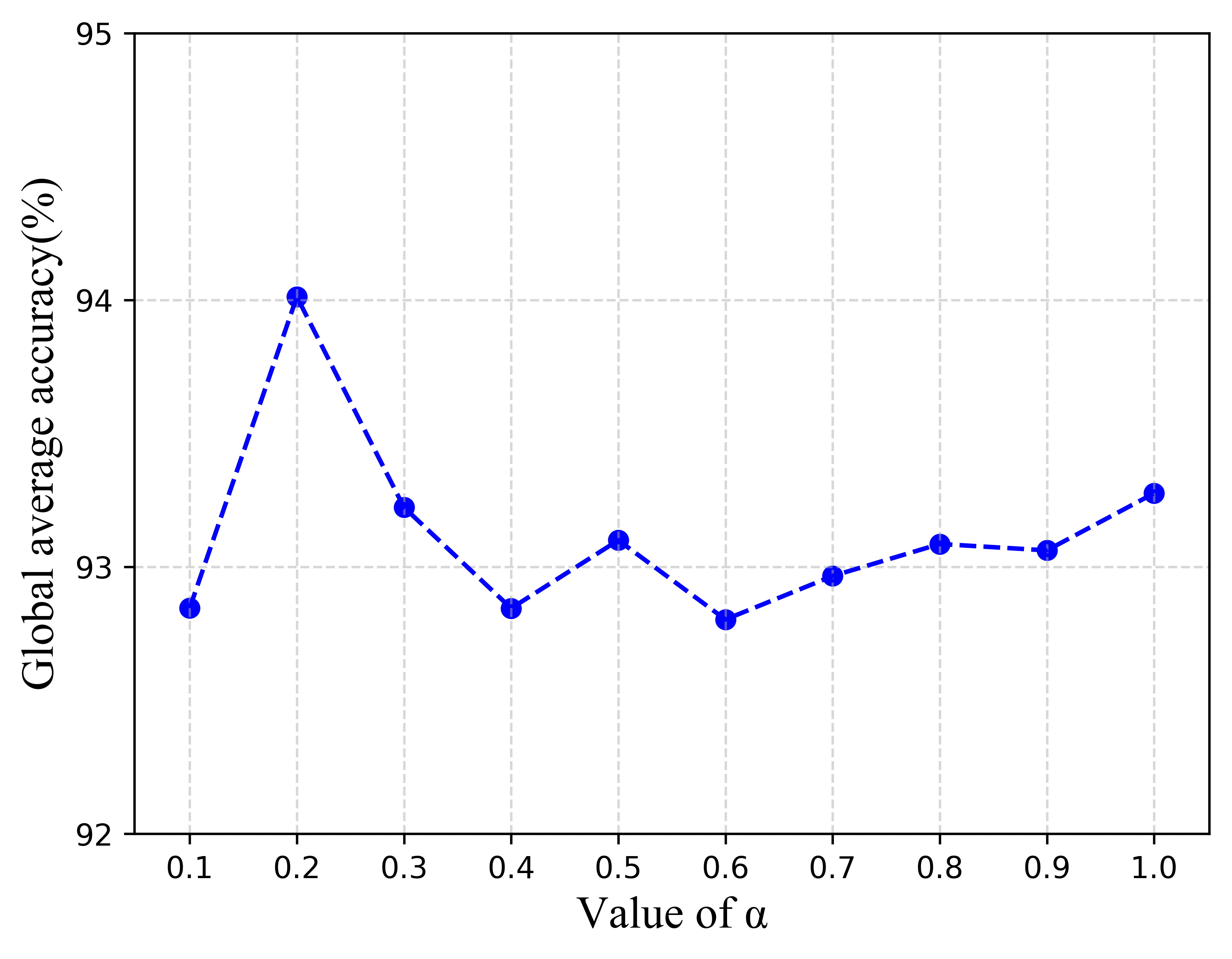}
	\caption{Global average accuracy of different values of $\alpha$ on UCF101.}
	\label{alpha_gaa}
\end{figure}

To determine the optimal value of $\alpha$, we analyze the global average accuracy for different values of $\alpha$ on UCF101, as shown in Fig.~\ref{alpha_gaa}, and find that $\alpha = 0.2$ works best. We emphasize two key differences from existing works: Firstly, the teacher generator generates a stable and accurate soft label, which prevents the largest logit from being significantly larger than the others and prevents the network from being disturbed by incorrect predictions from the teacher network. {Secondly, Label Smoothing (LS)~\cite{szegedy2016rethinking} is a special case of Eq.~(\ref{eq5}) if there is a linear constraint between ${\hat y}^{\kappa}$ and $p^{\kappa}$,  and $\alpha = 1$, i.e., ${L_{ls}}(x;{\theta _\kappa }{\text{)  =  }} - \sum\limits_{(x,y)} {((1 - \varepsilon )y + \frac{\varepsilon }{\mathcal{C}})log(q(x))} $ where $\varepsilon$ denotes a smoothing parameter.} Different from LS, our method provides an alternative to a teacher network and directs the network to review previous knowledge while performing the current task.

\noindent\textbf{Self-correction loss.} We put forward a self-correction loss to measure the difference between the network prediction and the label defined by $\chi_{\kappa}(x)$.

\begin{equation}
	{L_{sc}(x;{\theta _\kappa })} =  - \sum\limits_{x \in {\mathcal{P}_{\kappa - 1}}} \chi_{\kappa}(x) log(\frac{1}{{1 + \exp ( - {q(x)})}}).
\end{equation}

\DeclareGraphicsExtensions{.pdf,.jpeg,.png,.jpg}

\begin{figure}[htbp]
	\centering
	\includegraphics[scale=0.57]{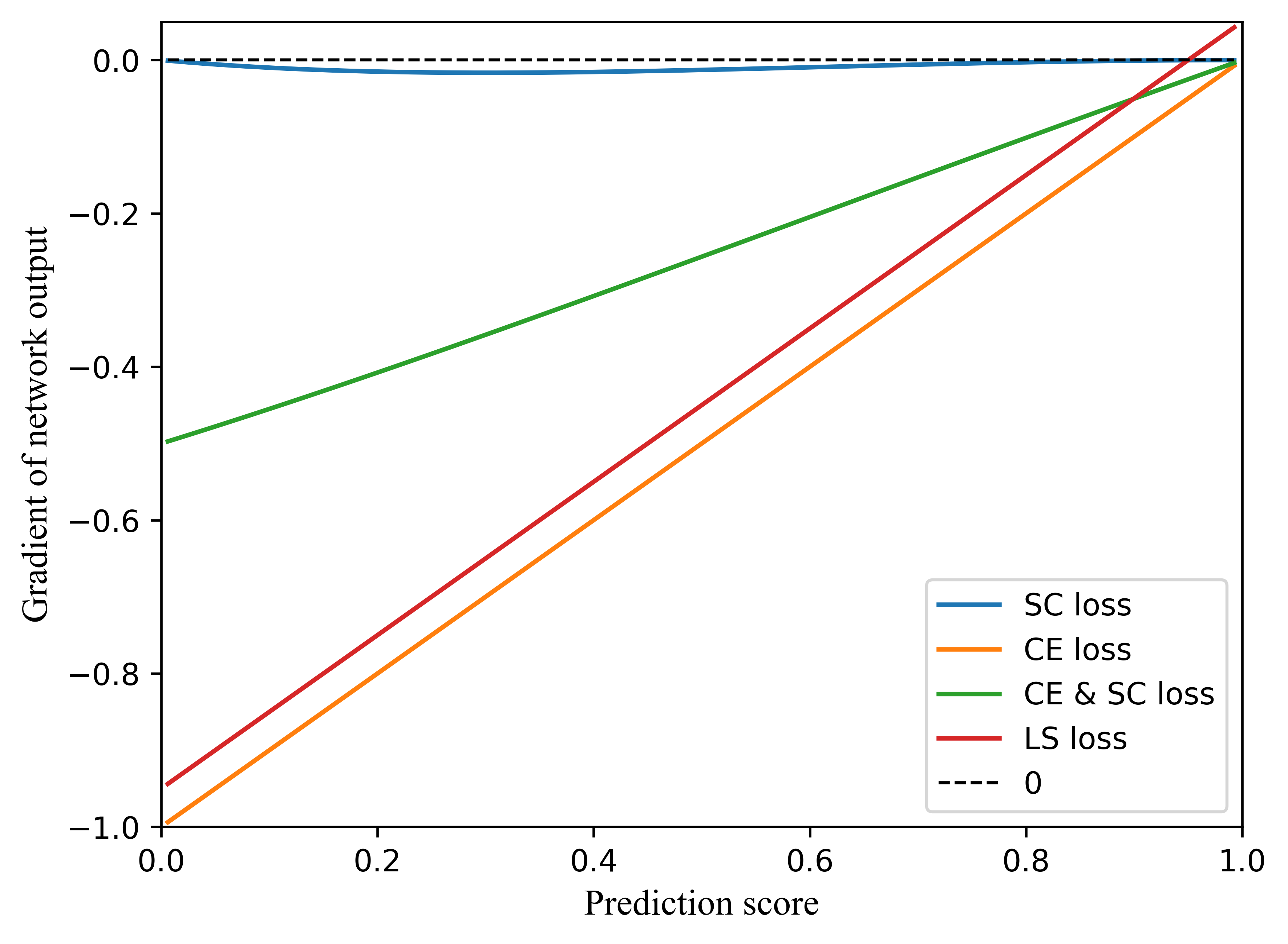}
	\caption{{The gradients of the network output for different losses. CE loss and LS loss denote the cross-entropy loss and label smoothing loss~\cite{szegedy2016rethinking}, respectively. CE \& SC loss comes from Eq. (3) ($\lambda = 0.5$). The probability $\varepsilon$ is set to 0.05 in LS loss. The gradient of the proposed SC Loss pays more attention to those memorized exemplars that are not completely forgotten, rather than treating different classification results linearly as observed in CE Loss and LS Loss.}}
	\label{sc_prof}
\end{figure}

We replace the distillation loss ${L_{kd}}(x;{\theta _\kappa })$ in Eq. (\ref{eq3}) with our proposed SC loss ${L_{sc}(x;{\theta _\kappa })}$ for knowledge review. {To gain a better understanding of the role of SC loss, we use binary classification as an example to demonstrate the impact of different loss gradients on network parameters in Fig.~\ref{sc_prof}. Specifically, we showcase the outcomes of gradient feedback from the final FC layer. To illuminate the gradient variations distinctly, we designate the label as 1, hold the output of the first FC node at 0.1, and investigate the influence of the second node's output—ranging from -5 to 5—on the ultimate gradient. }

{The experimental results underscore the positive impact of the proposed SC loss on network parameters, particularly when the predicted scores fall within the range of approximately 0.1 to 0.6. Beyond providing additional supervision signals for misclassified samples, the SC loss effectively steers the network to allocate increased attention to instances that have not been completely forgotten. In contrast, LS loss prioritizes addressing the influence of overconfidence on network parameter updates, generating a counteractive gradient in situations of excessively high confidence. The joint version of SC loss and CE loss not only circumvents the adverse gradient effect arising from excessive confidence but also guides the network to reevaluate existing memories.}




\DeclareGraphicsExtensions{.pdf,.jpeg,.png,.jpg}

\subsection{Unified Sampler}

As previously mentioned, the memory buffer has a limited capacity for storing samples. In the case of image-based incremental tasks, this is relatively straightforward to manage. However, when dealing with video tasks, it becomes more challenging since the total memory is not fixed due to the varying frame numbers of different videos. For this, a simple solution is to utilize the difference between consecutive frames to select key-frames~\cite{zhao2021video}. But the selected frame number is not fixed, and its effect depends on the selection of the threshold value, which usually needs to be adjusted for different scenes. \cite{zhi2021mgsampler} employed the image-level (or feature-level)  differences between consecutive RGB frames to capture motion-specific features and suppress the static background. {Subsequently, they utilized the cumulative distribution to determine a fixed number of key-frames for a single video clip. While the difference at the feature level exhibits a slight improvement compared to that at the image level, the latter operation is more direct and simple. Nevertheless, differences at the image level are prone to interference from both internal and external noise, such as lighting changes and motion blur. Additionally, despite the effective selection of key-frames through cumulative distribution, it lacks theoretical support. Drawing on insights from these two approaches, we introduce a theoretically interpretable unified sampler.} Formally, it is formulated as 


\begin{equation}\label{eq7}
	\begin{aligned}
		\frac{{\sum\limits_{n = j}^{i-1} {{{(|f({P_{n+1}}) - f({P_{n}})|)}^\gamma }} }}{{R(\zeta)}} \leqslant \beta,
	\end{aligned}
\end{equation}
where  $0 < j < i \leqslant \zeta $ ($\zeta$ is the total frame number), $ {P}_{n} $ represents the $n$-th frame in the current video sample, {$f( \cdot )$ serves as a mapping function for a video frame, with the objective of transforming the current RGB video frame into a representation tensor. Typical mapping functions include identity mapping, ${\ell _1}$-norm, and ${\ell _2}$-norm.}  $R(\zeta)={\sum\limits_{n = 1}^{\zeta-1} {{{(|f({{P}_{n+1}}) - f({{P}_{n}})|)}^\gamma }} }$, $0 < \gamma  \leqslant 1$ is a smoothing factor (0.5 by default). The above equation is derived from the key-frame selection formula in~\cite{zhao2021video}, where those exemplars are chosen based on the condition $\left| {{{{P}}_i} - {{{P}}_j}} \right| \leqslant \beta R(\zeta)$.  While this approach reduces the number of video frames, the frame number is not fixed and heavily dependent on the choice of $\beta$. To address this issue, we extend this method to a more general form through the following analysis:

\begin{equation}\label{eq8}
	\begin{gathered}
		|f({P_i}) - f({P_j}){|^{\gamma} } = |f({P_i}) - f({P_{i - 1}}) + f({P_{i - 1}}) - f({P_{i - 2}}) \hfill \\
		+  \cdots + f({P_{j + 1}}) - f({P_j}){|^{\gamma} } \leqslant \sum\limits_{n = j}^{i - 1} {{{(|f({P_{n + 1}}) - f({P_n})|)}^{\gamma} }}  \hfill \\
		\leqslant \beta R(\zeta) \hfill \\ 
	\end{gathered}
\end{equation}

After a simple transposition operation of Eq.~(\ref{eq8}), Eq.~(\ref{eq7}) can be obtained. The proposed method can utilize cumulative distribution for fixed-length frame selection based on a given threshold. Clearly, it is not difficult to find that~\cite{zhi2021mgsampler} is a special case of our method, and our analysis indicates the  theoretical rationality of this strategy. Our proposed sampler enables the selection of a fixed number of key-frames per video, which ensures a manageable memory overhead while maintaining network performance during knowledge replay.

\section{Experiments}\label{exper}
\subsection{Dataset}
We evaluate the proposed method on four public video  datasets. These datasets can be grouped into two types.
(1) Motion-related dataset:  Something-Something V1 (SS V1)~\cite{goyal2017something} comprises an extensive collection of video clips showcasing everyday actions involving commonplace objects. The dataset was collected by performing identical actions with various objects, with the objective of emphasizing motion rather than contextual factors such as objects or surroundings. The SS V1 dataset consists of 108,499 video clips spanning 174 action classes.  (2) Scene-related datasets: Kinetics-incremental, UCF101~\cite{soomro2012ucf101} and HMDB51~\cite{kuehne2011hmdb}. {Kinetics-incremental is a subset of the Kinetics dataset~\cite{carreira2017quo} that reduces the sample size of the original dataset by half, while still maintaining a minimum of 200 video clips per category. It has 145,123 video clips and includes 10-second videos sampled at 25 frames per second from YouTube, categorized into 400 action classes. Additionally, considering the challenges of accessing some video clips of the Kinetics dataset from the Internet, we will make this kinetics-incremental dataset publicly available for fair comparison.} Furthermore, UCF101 is a challenging dataset consisting of realistic action videos collected from YouTube, featuring 13,320 videos distributed among 101 action classes. The videos are classified into 25 groups, where each group may have some shared features, such as comparable backgrounds and viewpoints. In contrast, HMDB51 has roughly 7,000 videos distributed among 51 classes and is less sensitive to temporal relationships.


\subsection{Evaluation Protocol}

Incremental learning involves initially learning a base task from a small subset of the total classes and then gradually adding a certain number of new classes in each incremental step. For instance, we learn 40, 21, 11 and 6 classes in the base session for Kinetics-incremental, SS V1, UCF101 and HMDB51, respectively. The remaining classes are distributed evenly across nine groups for subsequent incremental sessions. Our report results are for 10-task unless explicitly stated otherwise. To maintain consistency with iCaRL~\cite{rebuffi2017icarl}, we select 20 exemplars for each old class using the herding method. Thus, we retain a maximum of 8000, 2020, 1020, and 3480 videos for kinetics-incremental,  SS V1, UCF101 and HMDB51, respectively. {For a fair comparison, we evaluate our method using three standard incremental learning metrics: Final Average Accuracy (Acc), Backward Forgetting (BWF), and Global Average Accuracy (GAA)~\cite{zhao2020maintaining,villa2022vclimb}. Acc measures the average classification accuracy across all learned tasks, and BWF quantifies the impact of the $i$th learned task on the performance of the previous tasks.} 
\begin{equation}
	Acc = \frac{1}{{\vec N}}\sum\limits_{i = 1}^{\vec N} {{A_{\vec N,i}}},
\end{equation}
{and}
\begin{equation}
	BWF = \frac{1}{{\vec N - 1}}\sum\limits_{i = 1}^{\vec N - 1} {{A_{i,i}} - } {A_{\vec N,i}},
\end{equation}
{where $\vec N$  denotes the overall count of observed tasks, and ${A_{\vec N,i}}$  denotes the accuracy achieved on the $i{\text{th}}$  task after training on task  $\vec N$.}

{GAA reflects the average Acc of all tasks to determine the average performance of each task.}
\begin{equation}
GAA = \frac{1}{{\vec M}}\sum\limits_{i = 1}^{\vec M} {A{cc_i}},
\end{equation}
{where $\vec M$  denotes the total number of tasks, and   $A{cc_i}$ denotes the accuracy of the $i{\text{th}}$ task.}


\begin{table*}[!htbp]
	\centering
	\caption{{Comparison with some SOTA methods in the 10-task setups of four action recognition benchmarks. 'Joint'  denotes training a model on all available data as the upper bound performance on each dataset. 'Mem.' denotes the memory size of the exemplars, where the numbers in this column represent the number of memory samples (1K = 1000) and the average frames of all samples in the corresponding dataset. Our method achieves superior performance across nearly all metrics, showcasing a substantial improvement compared to a strong baseline, vCLIMB\_TC.}}
	\scalebox{0.77}{
	\begin{tabular}{c|c|c|c|c|c|c|c|c|c|c|c|c|c|c|c|c}
		\hline
		\multirow{2}*{Method} & \multicolumn{4}{c|}{Kinetics-incremental} & \multicolumn{4}{c|}{Something-Something V1} & \multicolumn{4}{c|}{UCF101} & \multicolumn{4}{c}{HMDB51} \\
		\cline{2-17}
		&{Mem.}&{Acc$\uparrow$}&{BWF$\downarrow$}&{GAA$\uparrow$}&{Mem.}&{Acc$\uparrow$}&{BWF$\downarrow$}&{GAA$\uparrow$}&{Mem.}&{Acc$\uparrow$}&{BWF$\downarrow$}&{GAA$\uparrow$}&{Mem.}&{Acc$\uparrow$}&{BWF$\downarrow$}&{GAA$\uparrow$}\\
		\hline  
		{Fine-tuning}&-&7.38&78.08&23.38&-&5.75&59.42&15.94&-&9.90&99.57&29.27&-&9.31&91.88&24.85\\
		\cline{1-17}
		{LwF~\cite{li2017learning}}&-&7.69&78.22&23.00&-&5.62&58.96&15.91&-&9.90&98.80&30.63&-&8.90&93.25&25.39\\
		\cline{1-17}
		{iCaRL~\cite{rebuffi2017icarl}}&8K$\times$286f&27.68&25.26&49.03&3.48K$\times$49f&10.77&36.68&25.44&2.02K$\times$186f&80.61&19.08&92.53&1.02K$\times$185f&52.98&36.20&\textbf{70.93}\\
		\cline{1-17}
		{BiC~\cite{wu2019large}}&8K$\times$286f&25.08&34.10&46.64&3.48K$\times$49f&10.35&47.27&25.68&2.02K$\times$186f&79.32&22.08&91.66&1.02K$\times$185f&49.22&41.21&68.77\\
		\cline{1-17}
		{vCLIMB\_TC~\cite{villa2022vclimb}}&8K$\times$286f&32.81&27.63&50.00&3.48K$\times$49f&11.29&34.22&23.24&2.02K$\times$186f&75.82&25.02&90.23&1.02K$\times$185f&45.28&44.89&67.01\\
		\cline{1-17}
		{Ours}&8K$\times$16f&\textbf{38.65}&\textbf{22.48}&\textbf{51.86}&3.48K$\times$16f&\textbf{15.48}&\textbf{33.41}&\textbf{25.88}&2.02K$\times$16f&\textbf{85.53}&\textbf{14.77}&\textbf{94.01}&1.02K$\times$16f&\textbf{53.38}&\textbf{33.99}&67.73\\
		\cline{1-17}
		{Joint}&-&45.10&-&-&-&27.84&-&-&-&94.13&-&-&-&69.39&-&-\\ 
		\hline
	\end{tabular}
}
	\centering
	\label{sota}
\end{table*}

\begin{table*}[!htbp]
	\centering
	\caption{{Comparison with some SOTA methods in the 20-task setups of four action recognition benchmarks. With the exception of the incremental task setups, all other configurations mirror those of the 10-task setups, including network hyperparameter and memory size. Our network consistently achieves significant performance improvements on multiple metrics, surpassing other SOTA methods.}}
	\scalebox{0.8}{
	\begin{tabular}{c|c|c|c|c|c|c|c|c|c|c|c|c}
		\hline
		\multirow{2}*{Method} & \multicolumn{3}{c|}{Kinetics-incremental} & \multicolumn{3}{c|}{Something-Something V1} & \multicolumn{3}{c|}{UCF101} & \multicolumn{3}{c}{HMDB51} \\
		\cline{2-13}
		&{Acc$\uparrow$}&{BWF$\downarrow$}&{GAA$\uparrow$}&{Acc$\uparrow$}&{BWF$\downarrow$}&{GAA$\uparrow$}&{Acc$\uparrow$}&{BWF$\downarrow$}&{GAA$\uparrow$}&{Acc$\uparrow$}&{BWF$\downarrow$}&{GAA$\uparrow$}\\
		\hline  
		{Fine-tuning}&3.69&84.19&15.27&3.09&66.53&9.23&4.95&99.62&18.12&3.77&94.64&10.77\\
		\cline{1-13}
		{LwF~\cite{li2017learning}}&3.65&83.73&15.14&3.07&65.12&9.15&4.92&99.49&18.30&3.92&94.53&10.67\\
		\cline{1-13}
		{iCaRL~\cite{rebuffi2017icarl}}&22.71&\textbf{24.54}&46.73&7.82&41.21&20.99&72.30&26.56&89.42&44.71&42.10&\textbf{64.23}\\
		\cline{1-13}
		{BiC~\cite{wu2019large}}&19.84&35.53&45.06&8.30&42.93&21.28&69.95&25.04&86.90&44.17&36.39&61.16\\
		\cline{1-13}
		{vCLIMB\_TC~\cite{villa2022vclimb}}&28.43&36.35&47.47&8.90&43.13&19.87&72.59&27.87&88.96&41.86&44.89&54.03\\
		\cline{1-13}
		{Ours}&\textbf{35.38}&30.78&\textbf{50.40}&\textbf{13.52}&\textbf{37.17}&\textbf{22.19}&\textbf{82.32}&\textbf{16.44}&\textbf{91.45}&\textbf{50.80}&\textbf{30.50}&60.34\\
		\cline{1-13}
		{Joint}&45.10&-&-&27.84&-&-&94.13&-&-&69.39&-&-\\ 
		\hline
	\end{tabular}
	}
	\centering
	\label{sota_20}
\end{table*}


\subsection{Implementation Details}

Following the settings of~\cite{villa2022vclimb}, our backbone for this study is TSN \cite{wang2018temporal}, which employs the parameters of the ResNet-34 pretrained on ImageNet to initialize it. We adopt the same temporal data augmentation approach proposed in \cite{wang2018temporal} using $N = 8$ segments per video. We train the network using the Adam optimizer \cite{kingma2014adam} with a batch size of 12. The initial learning rate was set to $1\times{10}^{-3}$. To downsample the video frames, we resize the original resolution to 1.14 times the target resolution and then randomly crop the target resolution. Our unified sampler method enables us to select 16 exemplars for each video clip by $\beta = 1/16$. {Our code and Kinetics-incremental dataset will be released soon: \url{https://github.com/Tanyjiang/video_CIL}.}

\subsection{Comparison with SOTA Methods}

Here, we present a comparison of our proposed method against several state-of-the-art (SOTA) methods, including LwF~\cite{li2017learning}, iCARL~\cite{rebuffi2017icarl}, BiC~\cite{wu2019large}, and vCLIMB\_TC~\cite{villa2022vclimb}, on four challenging datasets. We also report the results of fine-tuning and joint-training methods, which are considered as lower and upper bounds for video incremental learning, respectively. As shown in Tables~\ref{sota} and~\ref{sota_20}, our method demonstrates superior results compared to the current SOTA methods across all evaluation metrics for both 10-task split and 20-task split. For 10-task split CIL, our method achieves much higher accuracy than the recent SOTA method vCLIMB\_TC by a large margin (17.80\% on Kinetics-incremental, 37.11\% on SS V1, 12.81\% on UCF101 and 17.89\% on HMDB51). {In comparison to the 10-task setups, the performance of all methods in the 20-task setups has experienced a decline. Notably, even in the face of the more challenging incremental configuration, our method consistently outperforms the accuracy of vCLIMB\_TC on all datasets in the 10-task version. This underscores the ability of our method to enhance the model's robustness across various incremental scenarios.  Despite achieving excellent performance, we would like to point out that the spatial resolution of input videos of our network is only half that of other methods. This highlights the efficiency and great potential of our method for class-incremental video learning.}

\begin{figure*}[htbp]
	\centering
	\includegraphics[scale=0.41]{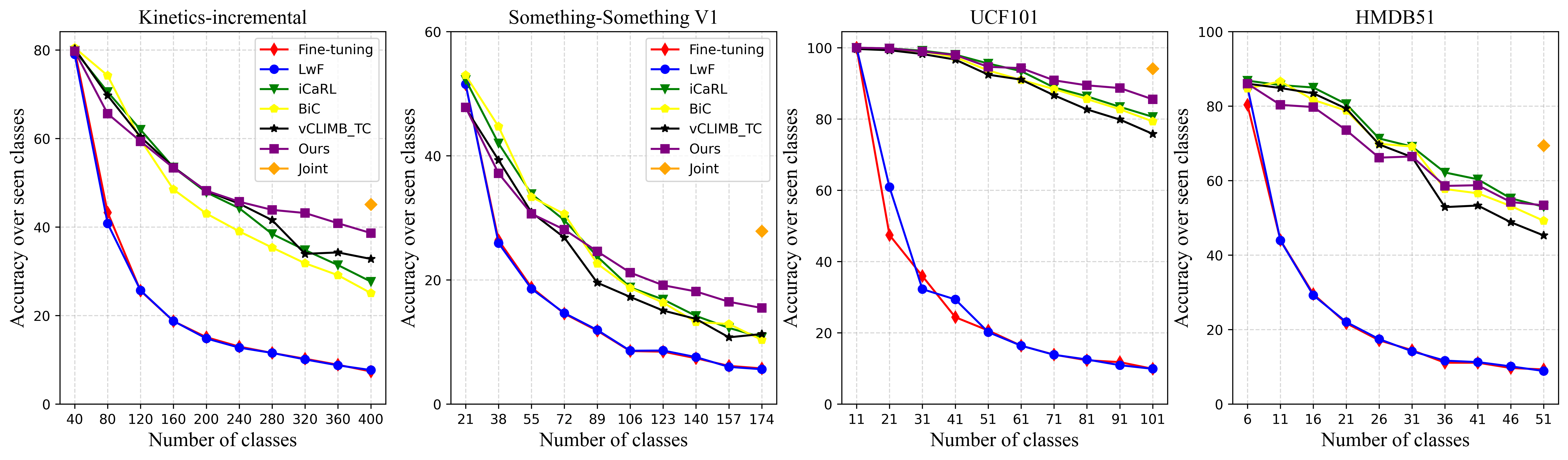}
	\caption{{Performance of different methods on Kinetics-incremental, Something-Something V1, UCF101 and HMDB51. The setups of all the methods are in the 10-task setups. As the number of learning tasks increases, our method exhibits robust stability and attains outstanding performance in the final stage.}}
	\label{different_dataset}
\end{figure*}

Moreover, we showcase the comparative performance of various methods in terms of their final average accuracy on four datasets, as depicted in Fig.~\ref{different_dataset}. Our method demonstrates superior accuracy on large datasets, namely Kinetics-incremental, SS V1 and UCF101, outperforming other methods, particularly in the latter tasks. On the smaller dataset HMDB51, our method initially lags behind iCaRL, BiC, and vCLIMB\_TC. However, as the tasks progress, our method displays more stability, ultimately surpassing iCaRL, the second-best performing method, in the final two tasks. This indicates the superior ability of our method to mitigate the catastrophic forgetting problem.

\subsection{ Ablation Study}
In this subsection, we conduct ablation studies to analyze the impact of different components of our method.

\begin{table*}[t]
	\centering
	\caption{{Ablation studies on different components of our method on Something-Something V1, UCF101 and HMDB51. Lr, Usam and T-agent denote low resolution, unified sampler, and teacher agent, respectively. As observed, low-resolution video sequences significantly degrade the performance of the network. The sequential integration of our proposed components, namely Usam and T-agent, contributes to substantial improvements in the final performance across three distinct types of datasets.}}
	\begin{tabular}{c|c|c|c|c|c|c|c|c|c|c|c}
		\hline
		\multirow{2}*{Lr}&\multirow{2}*{Usam}&\multirow{2}*{T-agent}&\multicolumn{3}{c|}{Something-Something V1}&\multicolumn{3}{c|}{UCF101}&\multicolumn{3}{c}{HMDB51}\\
		\cline{4-12}
		&&&{Acc$\uparrow$}&{BWF$\downarrow$}&{GAA$\uparrow$}&{Acc}&{BWF}&{GAA}&{Acc}&{BWF}&{GAA}\\
		\hline
		{\XSolidBrush}&{\XSolidBrush}&{\XSolidBrush}&11.29&34.22&23.24&75.82&25.02&90.23&45.28&44.89&67.01\\
		\cline{4-12}
		{\Checkmark}&{\XSolidBrush}&{\XSolidBrush}&8.23&33.24&21.37&67.32&29.97&87.21&43.65&47.90&65.90\\
		\cline{4-12}
		{\Checkmark}&{\Checkmark}&{\XSolidBrush}&14.48&31.68&24.72&69.88&30.65&87.26&48.81&37.21&64.53\\
		\cline{4-12}
		{\Checkmark}&{\Checkmark}&{\Checkmark}&15.48&33.41&25.88&85.53&14.77&94.01&53.38&33.99&67.73\\
		\hline		
	\end{tabular}
	\centering
	\label{components}
\end{table*}

\noindent \textbf{Effect of each component.} We have three key components in our method: Low resolution (Lr), Unified sampler (Usam), and Teacher agent (T-agent). Table~\ref{components} displays the results of these key components. The first row in this table shows the outcomes of vCLIMB\_TC, which we refer to as a \textit{\textbf{benchmark}} in the following text. As demonstrated, reducing the resolution causes the network's performance to suffer significantly, particularly on SS V1 and UCF101. This is understandable since lowering the resolution eliminates many video frame details, making it more challenging for the network to learn discriminative features. We consider this setup, i.e., vCLIMB\_TC with low-resolution input in the incremental stage, as a \textit{\textbf{baseline}}.

When adding unified sampler, the network's performance improves on all datasets, but more significantly on SS V1 and HMDB51. This implies that many video frames in the selected exemplars are redundant for knowledge replay, and thus eliminating them effectively will help to better recall old knowledge. Finally, the addition of SC loss enables our method to consistently enhance the network's performance on all datasets, especially with a 22.40\% improvement on UCF101. These outcomes are significantly better than those of the benchmark. Notably, we use lower resolution and smaller episodic memory in the incremental phase. Furthermore, our method does not rely on the teacher network during training, which can effectively decrease the computational and storage overhead while increasing the flexibility of incremental learning.

\begin{figure}[htbp]
	\includegraphics[scale=0.6]{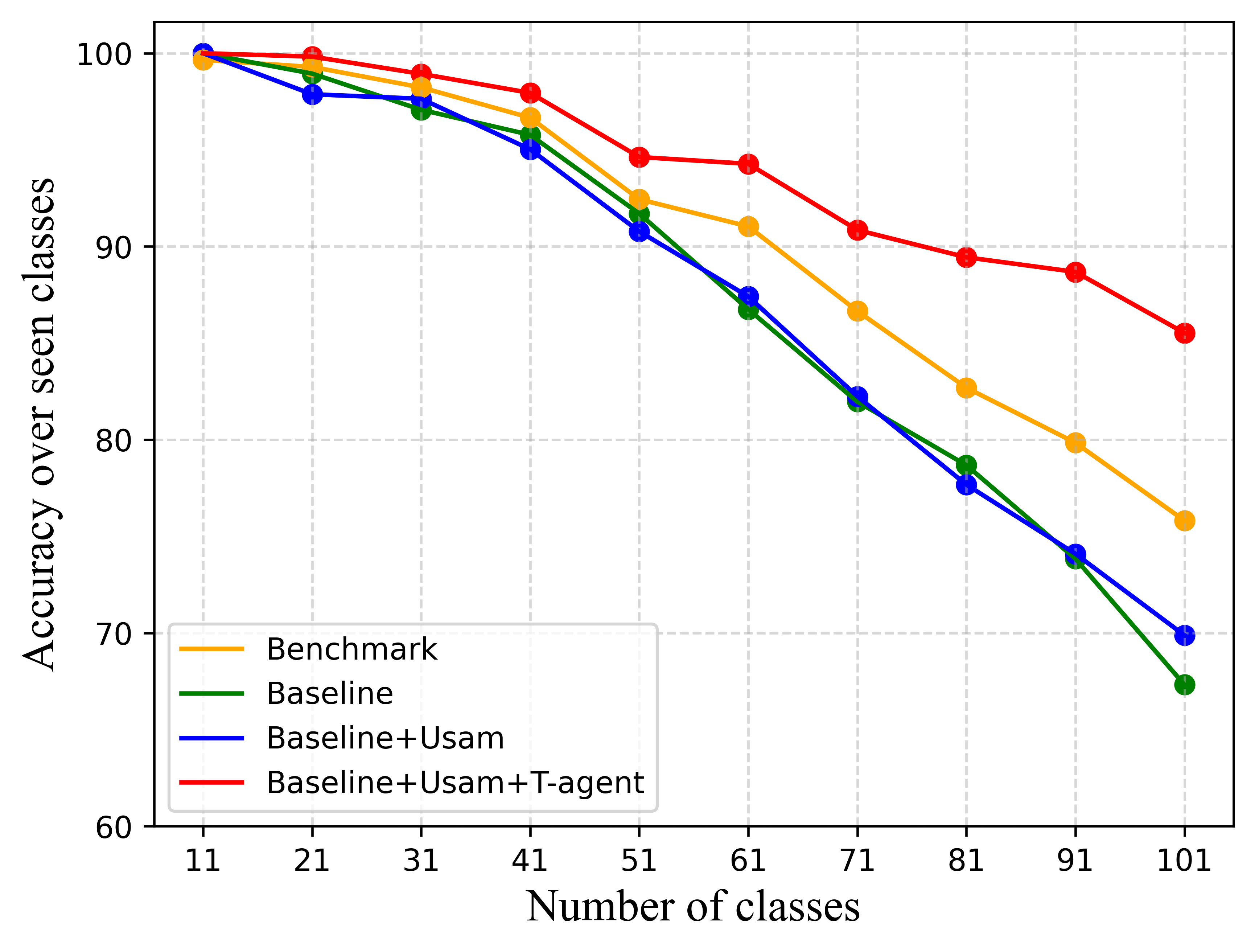}
	\caption{{Performance comparison of each component of our method on UCF101. During the incremental phase, the baseline experiences a notable decline in performance due to the reduction in video resolution when compared to the benchmark, vCLIMB\_TC. Usam demonstrates its capability to enhance model stability while concurrently reducing the number of memory video frames. Additionally, T-agent contributes significantly to the improvement of network performance.}}
	\label{component}
\end{figure}

To provide further evidence of the impact of different components, we present a visualization of their progressive performance across 10 tasks of UCF101 in Fig.~\ref{component}. The results indicate that our baseline, as well as the baseline with unified sampler (i.e., baseline + Usam),  underperforms benchmark. When T-agent is incorporated into the baseline with unified sampler, our method achieves a significant performance improvement, surpassing the benchmark by a large margin.  Moreover, the proposed method exhibits great stability in accuracy as the tasks progress, highlighting its robustness.

\noindent \textbf{Effect of spatial resolution.} Table~\ref{resolution} presents the performance at different spatial resolutions of video inputs on UCF101. As the resolution decreases, the network's performance also drops, but its computational complexity decreases significantly. As indicated in the table, the benchmark performance experienced a significant decline of 11.21\% when the resolution is reduced by half. In contrast, our method demonstrates remarkable resilience, with only a minor decrease of 3.41\% under the same resolution reduction. Notably, when the resolution of the input video frames is reduced by 1/4, our method still attains performance comparable to the baseline using the original resolution as input, namely benchmark. Simultaneously, the computational complexity of our method is significantly decreased, accounting for only 7.3\% of the FLOPs of the benchmark. By comparison, it is not difficult to find that the proposed strategy has a higher level of robustness against resolution reduction. Nevertheless, the performance of the network can be severely affected when the resolution is excessively low. In order to strike a balance between computational speed and performance, we choose $\delta$=0.5, i.e., we reduce the spatial resolution by half, as our experimental setting.



\begin{table*}[t]
	\centering
	\caption{{Comparison of recognition performance when the spatial resolution of video inputs varies on UCF101. As the reduction ratio of video resolution increases during incremental stages, the network demands fewer FLOPs, but there is a notable decrease in performance. Notably, even when reducing the resolution by 1/4 in the incremental stage, our method can achieve comparable results with the benchmark using original resolution as input.}}
	\begin{tabular}{c|c|c|c|c|c|c}
		\hline
		{Method}&{$\delta$}&{Resolution}&{FLOPs(G)$\downarrow$}&{Acc$\uparrow$}&{BWF$\downarrow$}&{GAA$\uparrow$}\\
		\cline{1-7}
		{Benchmark}&1&224$\times$224&58.85&75.82&25.02&90.23\\
		\cline{1-7}
		{Baseline}&0.5&112$\times$112&15.50&67.32&29.97&87.21\\
		\cline{1-7}
		\multirow{4}*{Ours}&1&224$\times$224&58.85&88.55&11.22&94.53\\
		&0.5&112$\times$112&15.50&85.53&14.77&94.01\\
		&0.25&56$\times$56&4.29&75.59&23.72&88.11\\
		&0.125&28$\times$28&1.14&59.05&37.26&79.17\\
		\hline		
	\end{tabular}
	\centering
	\label{resolution}
\end{table*}

\noindent \textbf{Effect of unified sampler.}  Table~\ref{sample} presents a comparison of various sampling methods with our method on SS V1 and UCF101. The first two methods involve selecting video frames using different distribution strategies. The uniform method divides the video uniformly into $N$ segments (16 by default) and then selects key-frames from these segments. On the other hand, the random method randomly chooses $N$ key-frames. The table shows that the uniform strategy achieves accuracy levels similar to the baseline on UCF101 while the random strategy slightly outperforms the baseline in terms of accuracy. But both falls slightly behind in terms of BWF and GAA. From an accuracy standpoint, utilizing all frames is not essential, but having redundant frames is beneficial in increasing sample diversity, which helps resist catastrophic forgetting and maintains high performance stability across multiple stages. 

Additionally, we  further compare our method with the other two sampling strategies, i.e., Dual-gra~\cite{zhao2021video} and Mgsampler~\cite{zhi2021mgsampler}. For the first one, we select top-$N$ samples with significant frame differences as key-frames to facilitate a fair comparison. The table reveals that Dual-gra is less effective than baseline on UCF101, likely due to the poor discriminative power of the selected key-frames resulting from substantial inter-frame noise. In contrast, Mgsampler demonstrates a more significant improvement in accuracy. This improvement can be attributed to the cumulative distribution, which better counteracts the effects of notable differences between individual frames. Compared with the aforementioned methods, our method excels at selecting fixed frames while achieving further improvements in accuracy. This is primarily due to the implementation of a smoothing factor, which effectively reduces the impact of noises on the inter-frame differences.

{In contrast to UCF101, the classification of SS V1 places a strong emphasis on capturing temporal information. The table highlights substantial improvements across three evaluation metrics when employing these sampling methods compared to the baseline. This suggests the presence of redundant information between frames in the video samples of this dataset, and these methods prove instrumental in enabling the model to extract valuable insights. Notably, our method consistently outperforms alternative methods, particularly in terms of accuracy and GAA. This underscores the efficacy of our proposed method.}



\begin{table*}[t]
	\centering
	\caption{{Performance comparison of different sampling methods.  "Baseline" denotes the preservation of every frame from the old representative samples. "Uniform" denotes that all frames of the video sample are divided into 16 segments, and the center frame from each segment is selected. For "Random", 16 frames are chosen randomly and saved from the entire set of frames. Our proposed method demonstrates outstanding performance on both motion-related and scene-related datasets.}}
	\begin{tabular}{c|c|c|c|c|c|c}
		\hline
		\multirow{2}*{Sampling Methods} & \multicolumn{3}{c|}{Something-Something V1} & \multicolumn{3}{c}{UCF101}\\
		\cline{2-7}
		&{Acc$\uparrow$}&{BWF$\downarrow$}&{GAA$\uparrow$}&{Acc$\uparrow$}&{BWF$\downarrow$}&{GAA$\uparrow$}\\
		\cline{1-7}
		{Baseline}&8.23&33.24&21.37&67.32&29.97&87.21\\
		\cline{1-7}
		{Uniform}&13.87&30.10&23.05&67.33&33.02&86.89\\
		\cline{1-7}
		{Random}&13.75&29.61&23.12&68.54&29.99&86.29\\
		\cline{1-7}
		{Dual-gra~\cite{zhao2021video}}&13.96&31.37&23.92&66.31&34.92&86.13\\
		\cline{1-7}
		{Mgsampler~\cite{zhi2021mgsampler}}&14.09&30.07&23.95&68.77&31.62&86.92\\
		\cline{1-7}
		{Usam(Ours)}&14.48&31.68&24.72&69.88&30.65&87.26\\
		\hline		
	\end{tabular}
	\centering
	\label{sample}
\end{table*}

\noindent \textbf{Effect of teacher agent.}  To have more insights into the teacher agent,  some ablation studies of this method on UCF101 are conducted as shown in Table \ref{scloss}. We have the following findings:

(1) Knowledge distillation is not an indispensable strategy for retaining old knowledge. Our study compares the effectiveness of various approaches against two baselines: one using knowledge distillation and the other removing mispredictions of teacher networks (baseline(EME)). In this comparison, both label smoothing~\cite{szegedy2016rethinking} and our proposed method demonstrate significant improvements in terms of Acc, BWF and GAA. Notably, our proposed method exhibits nearly double the performance in terms of BWF compared to baseline. This finding suggests that our method can effectively replace knowledge distillation during experience replay. An apparent advantage of these non-distillation methods is their ability to achieve remarkable results while consuming negligible computational resources, making them more promising alternatives.

(2) In comparison to the label smoothing mechanism, our method excels at providing stable regularization to the classification layer throughout multiple incremental stages, ultimately leading to superior results in the final stage. The evidence presented in Table~\ref{scloss} demonstrates that our methods surpasses the label smoothing method in terms of accuracy and average accuracy across various generators. This result highlights the stable and accurate labels generated by our strategy, coupled with the SC loss, can effectively guide the network to retain and recall old knowledge.

(3) A stable agent generator plays a crucial role in preserving old knowledge within the network. The proposed agent generator exhibits flexibility in accepting various inputs for label generation. Specifically, we can leverage the teacher network's output as input to the agent generator, without loss of generality. The results demonstrate that this strategy outperforms both the baseline with/without EME. Furthermore, we observed that ours(EME) performs better than ours(teacher), suggesting that the teacher model's elimination of hard samples contributes to knowledge retention. However, removing hard samples also reduces memory samples in the current stage. Before introducing non-teacher strategies, we introduce a baseline, i.e., ours(one-hot), which uses the ground-truth of old samples as the input of SC loss. It can be seen that the strategy is also better than baseline, indicating the effectiveness of the proposed loss. Next, we introduce a random proxy version, ours(rand), utilizing a random distribution as calibration input. Notably, this strategy outperforms the aforementioned methods. We argue that this method enhances label diversity and provides effective regularization for classification layers at each stage, contributing to improved overall performance.

Additionally, we explore the use of learnable parameters as input to the agent generator. To achieve this, we initialize the learnable parameters with small random values, which are then passed through a sigmoid nonlinear function to serve as the agent input. This method allows us to obtain more reasonable parameters through gradient back-up updates, resulting in more effective outputs. Through the table, we can get that this method has a good improvement in all three indicators compared with the random version. We notice that in the initial stage of network learning, these small random values lead to a more uniform label distribution after passing through the agent generator. Moreover, apart from favoring a specific non-true category with a larger value, it takes into account other categories equally, thereby preventing the network from becoming biased towards a particular category. Building upon this insight, we further freeze the parameters during the initial training stage. As shown in Table~\ref{scloss}, our method achieves notable improvements in network performance. In short, these results show that our proposed strategy can achieve effective performance improvement with minimal computational overhead.


(4) To evaluate the applicability of our proposed strategy, we extend its application to class-incremental image learning. Specifically, we incorporate the proposed method into two recent state-of-the-art methods, namely WA~\cite{zhao2020maintaining} and MEMO~\cite{zhou2022model}, using CIFAR-100 and Tinyimagenet datasets. In Table~\ref{imgCL}, we present a comparative analysis of the methods with T-agent against existing SOTA methods. The results clearly indicate that the method with T-agent can achieve performance improvements compared to the original methods. This provides further evidence of the effectiveness of our method and its potential applicability across different scenarios and datasets.

\begin{table}[t]
	\centering
	\caption{ Ablation studies of teacher agent on UCF101. Ours(teacher) denotes the calibration input characterized by teacher predictions;  Ours(EME) is an improved version of Ours(teacher) by eliminating the gradient back-propagation of misclassified exemplars; Ours(one-hot) denotes only using ground-truth of the exemplars as labels; Ours(rand) and Ours(para) denote the calibration input characterized by random vectors and learnable parameters with small random initialization, respectively. Ours is a parameter-frozen version of Ours(para).}
	\begin{tabular}{c|c|c|c}
		\hline
		{Method}&{Acc$\uparrow$}&{BWF$\downarrow$}&{GAA$\uparrow$}\\
		\cline{1-4}
		{Baseline}&67.32&29.97&87.21\\
		\cline{1-4}
		{Baseline(EME)}&68.67&31.38&87.71\\
		\cline{1-4}
		{Label smoothing}&83.48&16.95&92.73\\
		\cline{1-4}
		{Ours(teacher)}&83.62&17.08&92.94\\
		\cline{1-4}
		{Ours(EME)}&84.33&16.31&93.11\\
		\cline{1-4}
		Ours(one-hot) &82.74 &17.82 &92.89\\
		\cline{1-4}
		{Ours(rand)}&84.87&15.46&93.42\\
		\cline{1-4}
		{Ours(para)}&85.40&14.87 &93.60\\
		\cline{1-4}
		{Ours}&85.53&14.77&94.01\\
		\hline		
	\end{tabular}
	\centering
	\label{scloss}
\end{table}

\begin{table}[ht!]
	\centering
	\caption{{Performance comparison on CIFAR-100 and TinyImageNet. T-agent, a plug-and-play module, achieves new performance improvements by building upon SOTA models in image class incremental learning.}}
	\begin{tabular}{c|c|c|c|c}
		\hline
		\multirow{2}*{Method}&\multicolumn{2}{c|}{CIFAR-100}&\multicolumn{2}{c}{Tiny ImageNet}\\
		\cline{2-5}
		&5 tasks&10 tasks&5 tasks&10 tasks\\
		\cline{1-5}
		{LwF}&30.00&23.25&35.28&22.89\\
		\cline{2-5}
		{iCaRL}&54.23&49.52&44.03&35.85\\
		\cline{2-5}
		{BiC}&56.22&50.79&46.63&34.92\\
		\cline{1-5}
		{WA}&55.94&52.24&49.78&39.61\\
		\cline{1-5}
		{WA+T-agent}&57.09&52.85&49.92&40.62\\
		\cline{1-5}
		{MEMO}&60.95&57.52&52.45&46.36\\
		\cline{1-5}
		{MEMO+T-agent}&61.48&58.18&53.74&47.98\\
		\hline		
	\end{tabular}
	\centering
	\label{imgCL}
\end{table}

\noindent \textbf{Effect of memory size.}  The reduction in resolution greatly decreases the complexity of model computation and memory storage overhead, as evident above. This prompts us to investigate the impact of increasing the number of memory exemplars retained in memory on performance. In Table~\ref{memory}, we observe a significant improvement in our performance as the number of memory samples increases. Interestingly, our performance surpasses that of the joint training method on UCF101 when the number of memory samples is increased four-fold. This implies that we only need to store 80 exemplars per class such that our method achieves fully data-supervised performance with only 26\% FLOPs.

\begin{table*}[t]
	
	\centering
	\caption{{Performance comparison of the number of exemplars. $\times Num$ denotes the  multiple of the original stored exemplars (20 per class by default).  our method attains notable performance improvements across various datasets as the number of memorized video sequences increases. Notably, our method outperforms the joint training results on UCF101, when using four times the number of memory exemplars.}}
	\begin{tabular}{c|c|c|c|c|c|c}
		\hline
		{Dataset}&{Method}&{Mult}&{FLOPs(G)$\downarrow$}&{Acc$\uparrow$}&{BWF$\downarrow$}&{GAA$\uparrow$} \\
		\hline
		\multirow{4}*{SS V1}&Benchmark&{$\times 1$}&15.50&11.29&34.22&23.24\\
		\cline{2-7}
		~ &\multirow{3}*{Ours}&{$\times 1$}&\multirow{3}*{4.29}&15.48&33.41&25.88\\ 
		~ &&{$\times 2$}&&18.19&30.43&28.60\\
		~ &&{$\times 4$}&&21.08&26.18&31.31\\
		\hline
		\multirow{4}*{UCF101}&Benchmark&{$\times 1$}&58.85&75.82&25.02&90.23\\
		\cline{2-7}
		~ &\multirow{3}*{Ours}&{$\times 1$}&\multirow{3}*{15.50}&85.53&14.77&94.01\\
		~ &&{$\times 2$}&&91.36&8.49&96.24\\
		~ &&{$\times 4$}&&95.56&3.25&97.48\\
		\hline
		\multirow{4}*{HMDB51}&Benchmark&{$\times 1$}&58.85&45.28&44.89&67.01\\
		\cline{2-7}
		~ &\multirow{3}*{Ours}&{$\times 1$}&\multirow{3}*{15.50}&53.38&33.99&67.73\\
		~ &&{$\times 2$}&&59.46&27.69&71.81\\
		~ &&{$\times 4$}&&62.78&21.65&72.26\\
		
		\hline
	\end{tabular}
	\centering
	\label{memory}
\end{table*}

\begin{figure}[htbp]
	\centering
	\includegraphics[scale=0.6]{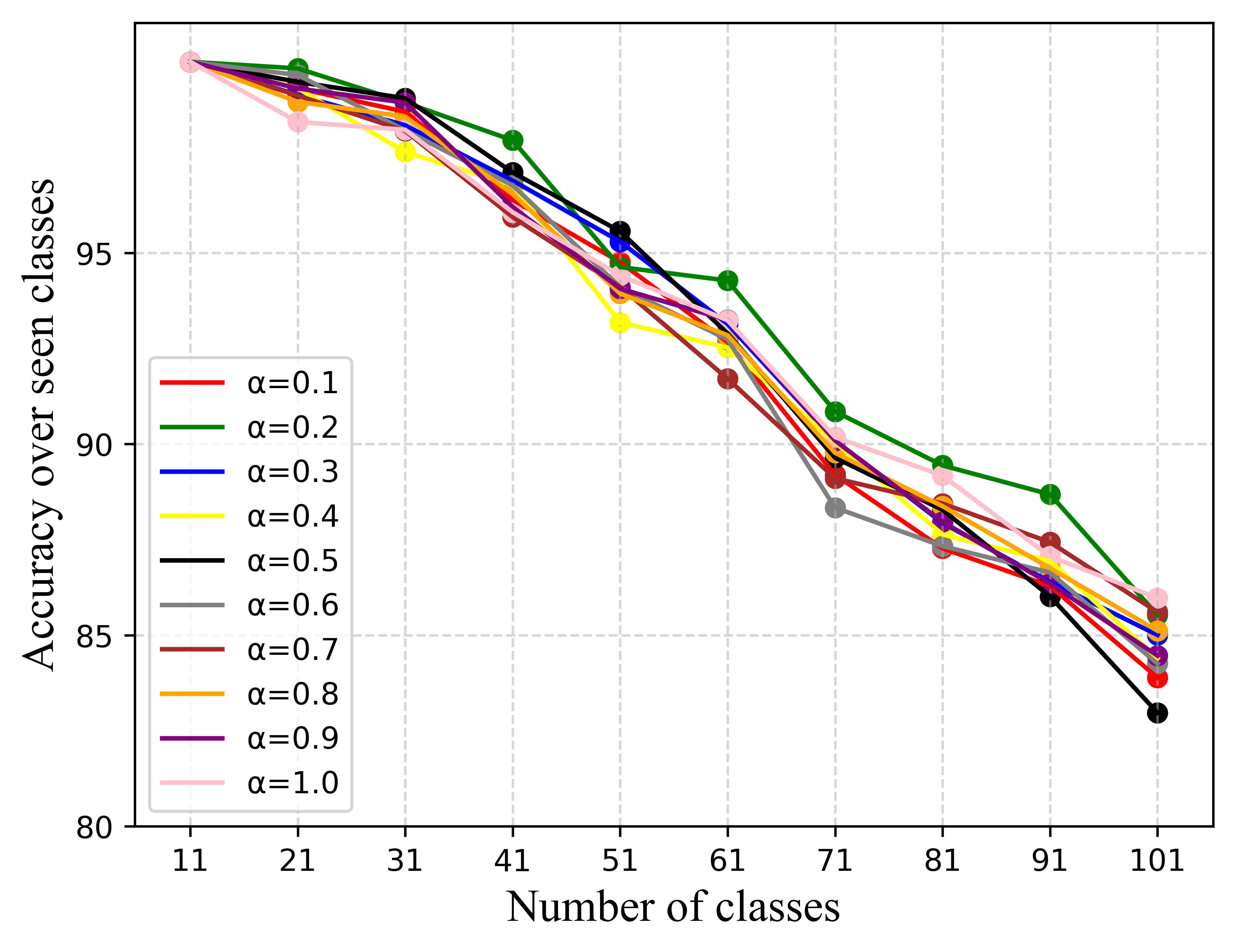}
	\caption{Final average accuracy of different values of $\alpha$ on UCF101.}
	\label{alpha_acc}
\end{figure}

\noindent \textbf{Effect of balance scalar $\alpha$.} Fig.~\ref{alpha_acc} presents the results for the average accuracy of different values of ${\alpha}$. The figure reveals that the network performs well with different ${\alpha}$ values. Although the performance is most pronounced in the final task when ${\alpha} = 1$, the global average accuracy at ${\alpha} = 0.2$ surpasses that of other values, as shown in Fig.~\ref{alpha_gaa}. This suggests that the network performance is more consistent across various tasks when ${\alpha} = 0.2$. As a result, we select ${\alpha} = 0.2$ as the default value for all experiments in this paper.

\DeclareGraphicsExtensions{.pdf,.jpeg,.png,.jpg}

\noindent \textbf{Effect of different backbones.} Table~\ref{backbone} compares three backbones with lower memory requirements, namely ResNet18, ResNet34~\cite{he2016deep} and MobileNet V2~\cite{sandler2018mobilenetv2}, on UCF 101. The table shows that our method achieves significant performance improvements for each backbone compared to the baseline. To a certain extent, it can be inferred that increasing the depth of the network or the number of its parameters could lead to improved performance. In this case, even with the use of lighter backbones such as ResNet18 and MobileNet V2, our method performs better than vCLIMB\_TC which utilizes a stronger backbone, ResNet34, as shown in Table~\ref{sota}.

\begin{table}[t]
	\centering
	\caption{{Performance comparison of different backbones on  UCF101. Backbone is an important factor affecting network performance. It is evident that among these relatively lightweight networks, ResNet34 stands out with the best performance.}}
	\begin{tabular}{c|c|c|c|c}
		\hline
		{Method}&{Backbone}&{Acc$\uparrow$}&{BWF$\downarrow$}&{GAA$\uparrow$}\\
		\cline{1-5}
		\multirow{3}*{Baseline}&{ResNet34}&67.32&29.97&87.21\\
		\cline{2-5}
		&{ResNet18}&65.62&34.22&84.31\\
		\cline{2-5}
		&{MobileNet V2}&60.66&37.42&82.71\\
		\cline{1-5}
		\multirow{3}*{Ours}&{ResNet34}&85.53&14.77&94.01\\
		\cline{2-5}
		&{ResNet18}&83.05&16.77&91.38\\
		\cline{2-5}
		&{MobileNet V2}&81.59&18.17&91.26\\
		\hline		
	\end{tabular}
	\centering
	\label{backbone}
\end{table}

\section{Conclusion}

In this study, we revisit the impact of knowledge distillation in the context of rehearsal-based video incremental tasks. Our experimental analysis reveals that the limited performance of the teacher network hinders the effective replay of old knowledge while requiring heavy computational overhead. To tackle these challenges, we present a novel video incremental learning framework that eliminates the need for knowledge distillation. Our framework employs a teacher agent to address the problem of catastrophic forgetting, featuring two key designs: the teacher generator and the self-correction loss. Firstly, we leverage a computationally efficient teacher generator to generate reliable labels. Secondly, we introduce a self-correction loss, serving as an effective regularization signal for retaining old knowledge. To reduce the memory buffer required for exemplars, we also introduce a unified sampler that efficiently mines key-frames. Extensive experiments show the efficacy of our proposed strategies, and our method outperforms some state-of-the-art methods by a large margin even when the spatial resolution of input video frames is reduced by half.


%

\appendices
%
%
%
%
%

\ifCLASSOPTIONcaptionsoff
  \newpage
\fi



%

{\small
	\bibliographystyle{ieee_fullname}
	\bibliography{cvbib}

\begin{thebibliography}{10}\itemsep=-1pt

\bibitem{aljundi2018memory}
Rahaf Aljundi, Francesca Babiloni, Mohamed Elhoseiny, Marcus Rohrbach, and
  Tinne Tuytelaars.
\newblock Memory aware synapses: Learning what (not) to forget.
\newblock In {\em The Proceedings of the European Conference on Computer
  Vision}, pages 139--154, 2018.

\bibitem{arnab2021vivit}
Anurag Arnab, Mostafa Dehghani, Georg Heigold, Chen Sun, Mario Lu{\v{c}}i{\'c},
  and Cordelia Schmid.
\newblock Vivit: A video vision transformer.
\newblock In {\em The Proceedings of the IEEE/CVF International Conference on
  Computer Vision}, pages 6836--6846, 2021.

\bibitem{bang2021rainbow}
Jihwan Bang, Heesu Kim, YoungJoon Yoo, Jung-Woo Ha, and Jonghyun Choi.
\newblock Rainbow memory: Continual learning with a memory of diverse samples.
\newblock In {\em The Proceedings of the IEEE/CVF Conference on Computer Vision
  and Pattern Recognition}, pages 8218--8227, 2021.

\bibitem{carreira2017quo}
Joao Carreira and Andrew Zisserman.
\newblock Quo vadis, action recognition? a new model and the kinetics dataset.
\newblock In {\em The Proceedings of the IEEE Conference on Computer Vision and
  Pattern Recognition}, pages 6299--6308, 2017.

\bibitem{cheraghian2021semantic}
Ali Cheraghian, Shafin Rahman, Pengfei Fang, Soumava~Kumar Roy, Lars Petersson,
  and Mehrtash Harandi.
\newblock Semantic-aware knowledge distillation for few-shot class-incremental
  learning.
\newblock In {\em The Proceedings of the IEEE/CVF Conference on Computer Vision
  and Pattern Recognition}, pages 2534--2543, 2021.

\bibitem{dosovitskiy2020image}
Alexey Dosovitskiy, Lucas Beyer, Alexander Kolesnikov, Dirk Weissenborn,
  Xiaohua Zhai, Thomas Unterthiner, Mostafa Dehghani, Matthias Minderer, Georg
  Heigold, Sylvain Gelly, et~al.
\newblock An image is worth 16x16 words: Transformers for image recognition at
  scale.
\newblock {\em arXiv preprint arXiv:2010.11929}, 2020.

\bibitem{douillard2020podnet}
Arthur Douillard, Matthieu Cord, Charles Ollion, Thomas Robert, and Eduardo
  Valle.
\newblock Podnet: Pooled outputs distillation for small-tasks incremental
  learning.
\newblock In {\em The Proceedings of the European Conference on Computer
  Vision}, pages 86--102. Springer, 2020.

\bibitem{goyal2017something}
Raghav Goyal, Samira Ebrahimi~Kahou, Vincent Michalski, Joanna Materzynska,
  Susanne Westphal, Heuna Kim, Valentin Haenel, Ingo Fruend, Peter Yianilos,
  Moritz Mueller-Freitag, et~al.
\newblock The" something something" video database for learning and evaluating
  visual common sense.
\newblock In {\em The Proceedings of the International Conference on Computer
  Vision}, pages 5842--5850, 2017.

\bibitem{he2016deep}
Kaiming He, Xiangyu Zhang, Shaoqing Ren, and Jian Sun.
\newblock Deep residual learning for image recognition.
\newblock In {\em Proceedings of the IEEE/CVF Conference on Computer Vision and
  Pattern Recognition}, pages 770--778, 2016.

\bibitem{hou2019learning}
Saihui Hou, Xinyu Pan, Chen~Change Loy, Zilei Wang, and Dahua Lin.
\newblock Learning a unified classifier incrementally via rebalancing.
\newblock In {\em The Proceedings of the IEEE/CVF Conference on Computer Vision
  and Pattern Recognition}, pages 831--839, 2019.

\bibitem{kang2022class}
Minsoo Kang, Jaeyoo Park, and Bohyung Han.
\newblock Class-incremental learning by knowledge distillation with adaptive
  feature consolidation.
\newblock In {\em The Proceedings of the IEEE/CVF Conference on Computer Vision
  and Pattern Recognition}, pages 16071--16080, 2022.

\bibitem{kingma2014adam}
Diederik~P Kingma and Jimmy Ba.
\newblock Adam: A method for stochastic optimization.
\newblock {\em arXiv preprint arXiv:1412.6980}, 2014.

\bibitem{kirkpatrick2017overcoming}
James Kirkpatrick, Razvan Pascanu, Neil Rabinowitz, Joel Veness, Guillaume
  Desjardins, Andrei~A Rusu, Kieran Milan, John Quan, Tiago Ramalho, Agnieszka
  Grabska-Barwinska, et~al.
\newblock Overcoming catastrophic forgetting in neural networks.
\newblock {\em The Proceedings of the National Academy of Sciences},
  114(13):3521--3526, 2017.

\bibitem{kuehne2011hmdb}
Hildegard Kuehne, Hueihan Jhuang, Est{\'\i}baliz Garrote, Tomaso Poggio, and
  Thomas Serre.
\newblock Hmdb: a large video database for human motion recognition.
\newblock In {\em The Proceedings of the International Conference on Computer
  Vision}, pages 2556--2563. IEEE, 2011.

\bibitem{li2020directional}
Xinyu Li, Bing Shuai, and Joseph Tighe.
\newblock Directional temporal modeling for action recognition.
\newblock In {\em The Proceedings of the European Conference on Computer
  Vision}, pages 275--291. Springer, 2020.

\bibitem{li2017learning}
Zhizhong Li and Derek Hoiem.
\newblock Learning without forgetting.
\newblock {\em IEEE Transactions on Pattern Analysis and Machine Intelligence},
  40(12):2935--2947, 2017.

\bibitem{liu2020mnemonics}
Yaoyao Liu, Yuting Su, An-An Liu, Bernt Schiele, and Qianru Sun.
\newblock Mnemonics training: Multi-class incremental learning without
  forgetting.
\newblock In {\em The Proceedings of the IEEE/CVF Conference on Computer Vision
  and Pattern Recognition}, pages 12245--12254, 2020.

\bibitem{liu2022video}
Ze Liu, Jia Ning, Yue Cao, Yixuan Wei, Zheng Zhang, Stephen Lin, and Han Hu.
\newblock Video swin transformer.
\newblock In {\em The Proceedings of the IEEE/CVF Conference on Computer Vision
  and Pattern Recognition}, pages 3202--3211, 2022.

\bibitem{park2021class}
Jaeyoo Park, Minsoo Kang, and Bohyung Han.
\newblock Class-incremental learning for action recognition in videos.
\newblock In {\em Proceedings of the IEEE/CVF International Conference on
  Computer Vision}, pages 13698--13707, 2021.

\bibitem{rebuffi2017icarl}
Sylvestre-Alvise Rebuffi, Alexander Kolesnikov, Georg Sperl, and Christoph~H
  Lampert.
\newblock Icarl: Incremental classifier and representation learning.
\newblock In {\em The Proceedings of the IEEE Conference on Computer Vision and
  Pattern Recognition}, pages 2001--2010, 2017.

\bibitem{sandler2018mobilenetv2}
Mark Sandler, Andrew Howard, Menglong Zhu, Andrey Zhmoginov, and Liang-Chieh
  Chen.
\newblock Mobilenetv2: Inverted residuals and linear bottlenecks.
\newblock In {\em Proceedings of the IEEE/CVF Conference on Computer Vision and
  Pattern Recognition}, pages 4510--4520, 2018.

\bibitem{soomro2012ucf101}
Khurram Soomro, Amir~Roshan Zamir, and Mubarak Shah.
\newblock Ucf101: A dataset of 101 human actions classes from videos in the
  wild.
\newblock {\em arXiv preprint arXiv:1212.0402}, 2012.

\bibitem{szegedy2016rethinking}
Christian Szegedy, Vincent Vanhoucke, Sergey Ioffe, Jon Shlens, and Zbigniew
  Wojna.
\newblock Rethinking the inception architecture for computer vision.
\newblock In {\em The Proceedings of the IEEE Conference on Computer Vision and
  Pattern Recognition}, pages 2818--2826, 2016.

\bibitem{villa2023pivot}
Andr{\'e}s Villa, Juan~Le{\'o}n Alc{\'a}zar, Motasem Alfarra, Kumail Alhamoud,
  Julio Hurtado, Fabian~Caba Heilbron, Alvaro Soto, and Bernard Ghanem.
\newblock Pivot: Prompting for video continual learning.
\newblock In {\em The Proceedings of the IEEE/CVF Conference on Computer Vision
  and Pattern Recognition}, pages 24214--24223, 2023.

\bibitem{villa2022vclimb}
Andr{\'e}s Villa, Kumail Alhamoud, Victor Escorcia, Fabian Caba, Juan~Le{\'o}n
  Alc{\'a}zar, and Bernard Ghanem.
\newblock vclimb: A novel video class incremental learning benchmark.
\newblock In {\em The Proceedings of the IEEE/CVF Conference on Computer Vision
  and Pattern Recognition}, pages 19035--19044, 2022.

\bibitem{wang2021tdn}
Limin Wang, Zhan Tong, Bin Ji, and Gangshan Wu.
\newblock Tdn: Temporal difference networks for efficient action recognition.
\newblock In {\em The Proceedings of the IEEE/CVF Conference on Computer Vision
  and Pattern Recognition}, pages 1895--1904, 2021.

\bibitem{wang2018temporal}
Limin Wang, Yuanjun Xiong, Zhe Wang, Yu Qiao, Dahua Lin, Xiaoou Tang, and Luc
  Van~Gool.
\newblock Temporal segment networks for action recognition in videos.
\newblock {\em IEEE Transactions on Pattern Analysis and Machine Intelligence},
  41(11):2740--2755, 2018.

\bibitem{wang2021action}
Zhengwei Wang, Qi She, and Aljosa Smolic.
\newblock Action-net: Multipath excitation for action recognition.
\newblock In {\em The Proceedings of the IEEE/CVF Conference on Computer Vision
  and Pattern Recognition}, pages 13214--13223, 2021.

\bibitem{wu2019large}
Yue Wu, Yinpeng Chen, Lijuan Wang, Yuancheng Ye, Zicheng Liu, Yandong Guo, and
  Yun Fu.
\newblock Large scale incremental learning.
\newblock In {\em The Proceedings of the IEEE/CVF Conference on Computer Vision
  and Pattern Recognition}, pages 374--382, 2019.

\bibitem{yan2021dynamically}
Shipeng Yan, Jiangwei Xie, and Xuming He.
\newblock Der: Dynamically expandable representation for class incremental
  learning.
\newblock In {\em The Proceedings of the IEEE/CVF Conference on Computer Vision
  and Pattern Recognition}, pages 3014--3023, 2021.

\bibitem{yang2022recurring}
Jiewen Yang, Xingbo Dong, Liujun Liu, Chao Zhang, Jiajun Shen, and Dahai Yu.
\newblock Recurring the transformer for video action recognition.
\newblock In {\em The Proceedings of the IEEE/CVF Conference on Computer Vision
  and Pattern Recognition}, pages 14063--14073, 2022.

\bibitem{yuan2020revisiting}
Li Yuan, Francis~EH Tay, Guilin Li, Tao Wang, and Jiashi Feng.
\newblock Revisiting knowledge distillation via label smoothing regularization.
\newblock In {\em The Proceedings of the IEEE/CVF Conference on Computer Vision
  and Pattern Recognition}, pages 3903--3911, 2020.

\bibitem{zhang2020class}
Junting Zhang, Jie Zhang, Shalini Ghosh, Dawei Li, Serafettin Tasci, Larry
  Heck, Heming Zhang, and C-C~Jay Kuo.
\newblock Class-incremental learning via deep model consolidation.
\newblock In {\em The Proceedings of the IEEE/CVF Winter Conference on
  Applications of Computer Vision}, pages 1131--1140, 2020.

\bibitem{zhao2020maintaining}
Bowen Zhao, Xi Xiao, Guojun Gan, Bin Zhang, and Shu-Tao Xia.
\newblock Maintaining discrimination and fairness in class incremental
  learning.
\newblock In {\em The Proceedings of the IEEE/CVF Conference on Computer Vision
  and Pattern Recognition}, pages 13208--13217, 2020.

\bibitem{zhao2021video}
Hanbin Zhao, Xin Qin, Shihao Su, Yongjian Fu, Zibo Lin, and Xi Li.
\newblock When video classification meets incremental classes.
\newblock In {\em The Proceedings of the ACM International Conference on
  Multimedia}, pages 880--889, 2021.

\bibitem{zhi2021mgsampler}
Yuan Zhi, Zhan Tong, Limin Wang, and Gangshan Wu.
\newblock Mgsampler: An explainable sampling strategy for video action
  recognition.
\newblock In {\em The Proceedings of the IEEE/CVF International Conference on
  Computer Vision}, pages 1513--1522, 2021.

\bibitem{zhou2022forward}
Da-Wei Zhou, Fu-Yun Wang, Han-Jia Ye, Liang Ma, Shiliang Pu, and De-Chuan Zhan.
\newblock Forward compatible few-shot class-incremental learning.
\newblock In {\em The Proceedings of the IEEE/CVF Conference on Computer Vision
  and Pattern Recognition}, pages 9046--9056, 2022.

\bibitem{zhou2022model}
Da-Wei Zhou, Qi-Wei Wang, Han-Jia Ye, and De-Chuan Zhan.
\newblock A model or 603 exemplars: Towards memory-efficient class-incremental
  learning.
\newblock {\em arXiv preprint arXiv:2205.13218}, 2022.

\bibitem{zhu2022self}
Kai Zhu, Wei Zhai, Yang Cao, Jiebo Luo, and Zheng-Jun Zha.
\newblock Self-sustaining representation expansion for non-exemplar
  class-incremental learning.
\newblock In {\em The Proceedings of the IEEE/CVF Conference on Computer Vision
  and Pattern Recognition}, pages 9296--9305, 2022.

\end{thebibliography}
}

%
%

%
%
%
%
%




\end{document}